\documentclass{article}
\usepackage[preprint]{neurips_2025} 
\usepackage[T1]{fontenc}
\usepackage{mathptmx} 
\DeclareFontShape{T1}{ptm}{m}{scit}{<-> ssub * ptm/m/sc}{}
\usepackage{algorithm}   
\usepackage{algpseudocode} 
\usepackage{amsmath}      
\usepackage{multirow}  
\usepackage{graphicx}  
\usepackage{booktabs}  
\usepackage[utf8]{inputenc} 
\usepackage{hyperref}       
\usepackage{url}            
\usepackage{booktabs}       
\usepackage{amsfonts}       
\usepackage{nicefrac}       
\usepackage{microtype}      
\usepackage{bbm}
\usepackage{epsfig}
\usepackage{amssymb}
\usepackage{caption}
\usepackage{subcaption}
\usepackage{bbding}
\usepackage{pifont}
\usepackage{soul}
\usepackage[accsupp]{axessibility}

\usepackage{tabularx}
\usepackage{adjustbox}

\usepackage{tcolorbox}

\addtocontents{toc}{\protect\setcounter{tocdepth}{0}} 
\usepackage{times}
\usepackage{latexsym}
\usepackage{inconsolata}
\usepackage{algpseudocode} 
\usepackage{graphicx}  
\usepackage{booktabs}  
\usepackage[T1]{fontenc}    
\usepackage{hyperref}       
\PassOptionsToPackage{table,xcdraw,dvipsnames}{xcolor}
\usepackage{amsfonts}       
\usepackage{nicefrac}       

\usepackage{marvosym}
\usepackage{epsfig}
\usepackage{amssymb}
\usepackage{caption}
\usepackage{subcaption}
\usepackage{bbding}
\usepackage{pifont}
\usepackage{soul}
\usepackage[accsupp]{axessibility}
\usepackage{tabularx}
\usepackage{adjustbox}
\usepackage{tcolorbox}
\addtocontents{toc}{\protect\setcounter{tocdepth}{0}}
\usepackage{colortbl}               
\usepackage{array}                  
\usepackage{makecell}               
\usepackage{geometry}
\usepackage{arydshln}
\usepackage{tabularx}

\definecolor{citecolor}{rgb}{0,0.443,0.737} 
\definecolor{linkcolor}{rgb}{0.956,0.298,0.235} 
\hypersetup{
    colorlinks=true,%
    citecolor=citecolor,%
    filecolor=citecolor,%
    linkcolor=linkcolor,%
    urlcolor=magenta
}

\usepackage{xspace}
\newcommand{\name}{\textsc{TlDr}\xspace}

\newtcolorbox{prompt}[1]{
    left=4mm,
    right=4mm,
    top=1mm,
    bottom=1mm,
    boxsep=0mm,
    rounded corners,
    title=#1,
    fonttitle=\normalsize\bfseries, 
    fontupper=\normalsize\linespread{0.7}\fontfamily{lmr}\selectfont,
}

\makeatletter
\def\adl@drawdash{\vrule\@height\adl@hlnth\@depth\adl@dshdp\@width\adl@dashlen\hskip\adl@dashgap}
\makeatother

\title{\textit{TL;DR}: Too Long, Do Re-weighting for Efficient LLM Reasoning Compression}

\author{
  Zhong-Zhi Li$^{\chi\pi}$\thanks{Equal contribution. Work done during internships at Microsoft.}~, Xiao Liang$^{\rho\gamma\,*}$, Zihao Tang$^{\phi}$, Lei Ji$^{\phi}$, Peijie Wang$^{\chi\pi}$, \\
  \textbf{Haotian Xu$^\gamma$, Xing W$^\pi$, Haizhen Huang$^{\phi}$, Weiwei Deng$^{\phi}$}, \\
  \textbf{Yeyun Gong$^{\phi}$, Zhijiang Guo$^{\theta\beta}$\thanks{Correspondence to Zhijiang Guo, Xiao Liu and Cheng-Lin Liu.
  \Letter: \texttt{zhijiangguo@hkust-gz.edu.cn}; \texttt{xiaoliu2@microsoft.com}; \texttt{ liucl@nlpr.ia.ac.cn}.}, Xiao Liu$^{\phi\dagger}$, Fei Yin$^{\chi\pi}$, Cheng-Lin Liu$^{\chi\pi\dagger}$}  \\
$^\chi$School of Artificial Intelligence, Chinese Academy of Sciences\\
$^\pi$Institute of Automation, Chinese Academy of Sciences\\
$^\rho$University of California, Los Angeles\quad
$^\gamma$Tsinghua University\\
$^\phi$Microsoft\quad
$^\beta$Hong Kong University of Science and Technology\\
$^\theta$Hong Kong University of Science and Technology (Guangzhou)\\
\textit{\url{https://github.com/zzli2022/TLDR}}
}
\begin{document}
\maketitle

\begin{abstract}
\vspace{-1.5pt}
Large Language Models (LLMs) have recently achieved remarkable progress by leveraging Reinforcement Learning and extended Chain-of-Thought (CoT) techniques. However, the challenge of performing efficient language reasoning—especially during inference with extremely long outputs—has drawn increasing attention from the research community. In this work, we propose a dynamic ratio-based training pipeline that does not rely on sophisticated data annotations or interpolation between multiple models. We continuously balance the weights between the model's System-1 and System-2 data to eliminate redundant reasoning processes while preserving the model's reasoning capability. We validate our approach across models on DeepSeek-R1-Distill-7B and DeepSeek-R1-Distill-14B and on a diverse set of benchmarks with varying difficulty levels. Our method significantly reduces the number of output tokens by nearly 40\% while maintaining the accuracy of the reasoning. Our code and data will be available soon.

\centering
\includegraphics[width=0.8\linewidth]{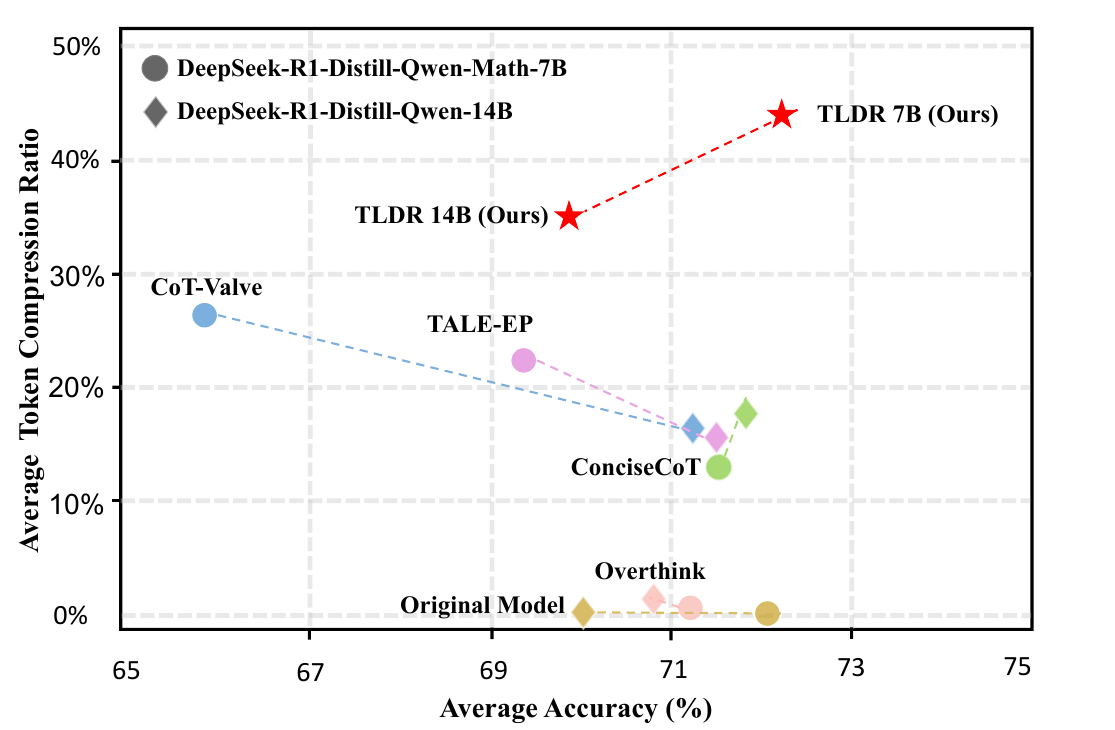} \\
\vspace{0.5em}
{\small Comparison of TLDR and baseline models in terms of average accuracy and token compression ratio. Higher values on both axes indicate better performance.}
\label{fig:tokens}
\vspace{-2mm}
\end{abstract}

\section{Introduction}

\begin{quote}
    \textit{``He that can have patience can have what he will.'' — Benjamin Franklin}
\end{quote}

Recent efforts have developed reasoning-oriented Large Language Models (LLMs) capable of solving complex tasks. These models progressed from System 1 to System 2 paradigms \citep{yu2024distilling21, li_system_2_reason}.  System 1 implementations, such as GPT-4o \citep{GPT4o}, LLaMA-3 \citep{llama3}, leverage rapid intuitive processing for immediate responses but struggle with complex 
reasoning tasks. In contrast, System 2 architectures such as DeepSeek-R1~\citep{Deepseek-R1} are fine-tuned with extended thinking chains to promote deliberate analysis through iterative self-assessment, error mitigation, and verification. However, reasoning-oriented LLMs, which employ system-2 reasoning, tend to engage in excessive deliberation even for simple problems. This results in unnecessary exploration and planning, ultimately impairing their efficiency and practicality.

Direct approaches aim to addressing the issues of cognitive redundancy and excessive deliberation within reasoning LLMs.  \textit{Training-free} methods~\cite{CoD, yao2025activation, TALE-EP} include some that control the internal states of the model during reasoning through prompts or confidence-based techniques to compress the model. Alternatively, the pathway, exemplified by model merging, involves intervening in the parameters of the reasoning LLM to produce relatively concise solutions. \textit{Training-based} methods primarily focus on sampling and synthesizing relatively concise reasoning paths on specified problem sets through various strategies \cite{TokenSkip, TOPS, CoT-Valve}. These methods involve performing reinforcement learning \cite{SimPO, thinkpruner, O1-Pruner, L1} or supervised fine-tuning (SFT) \cite{Overthink} on reasoning LLMs, enabling the model to learn to generate more concise yet still correct reasoning paths.

Such methods typically require careful collection of problems and precise control of the data ratio for different lengths to achieve good results, leading to a complex process of parameter tuning and data construction. For example, TOPS \cite{TOPS} requires pre-processing steps to manually label SFT data to construct length-sensitive models, while CoT-Valve \cite{CoT-Valve} generates data by creating intermediate models through model interpolation for sampling. This construction process is often tedious \cite{TOPS}, computationally expensive \cite{L1}, or difficult to control for quality \cite{CoT-Valve}. 

\begin{figure}[H]
    \centering
    \begin{tcolorbox}[
        width=\textwidth, 
        colframe=orange!75!black, 
        colback=orange!10,        
        coltitle=black,           
        title=Demystifying Short/Long CoT Mixture in LLM Thinking Compression,
        fonttitle=\bfseries       
    ]
    \small 
    \centering
    \begin{itemize}
        \item \textbf{System-1 data (Short CoT on GSM8K-like easy problems) reduces reasoning redundancy on all problem levels.} (Section \ref{pre_section:takeaway1_2}) Short CoT reduces reasoning redundancy on simple questions and demonstrates generalization across varying levels of problem difficulty. 
        \item \textbf{System-2 data (Long CoT only on s1-like difficult problems) helps maintain performance.} (Section \ref{pre_section:takeaway1_2}) Incorporating a small proportion of Long CoT, particularly on challenging problems, can mitigate the accuracy degradation introduced by short CoT, while long CoT on simple questions doesn’t help much.
        \item \textbf{Dynamic re-weighting of System-1/2 data builds effcient LLM Reasoning Compression.} (Section \ref{section:dynamic_takeaway}) Driven by a simple intuition, we design a dynamic reweighting algorithm for system-1/2 data, achieving strong performance in LLM reasoning compression. 
    \end{itemize}
    \end{tcolorbox}
    \vspace{-4mm}
\end{figure}

We investigate the effects of mixing short CoT and long CoT data on compressing reasoning. Our findings suggest that long CoT and short CoT induce divergent optimization directions in the model's reasoning behavior. Increasing the proportion of short CoT encourages more concise reasoning patterns, but may lead to a decline in reasoning accuracy. In contrast, raising the proportion of long CoT helps preserve reasoning performance on complex tasks, though at the expense of reduced compression efficiency. This naturally raises the question: \textit{Can we identify an \textbf{optimal Long-to-Short data mixture} that strikes the best trade-off—\textbf{maximizing reasoning efficiency} while \textbf{maintaining accuracy?}}

We base our approach on an intuitive motivation: when a model is thinking too long, it should re-weight more intuitive reasoning paths to simplify the thinking process. Conversely, when the thinking is too direct, it should incorporate more slow-thinking reasoning chains to encourage deeper contemplation. We propose a dynamic \textbf{T}hinking \textbf{L}ength \textbf{D}ata \textbf{R}e-Weighting method (\textbf{\name}), which dynamically balances the model’s complex reasoning using long CoT and efficient reasoning using short CoT data, enabling the model to eliminate redundant cognitive processes. First, we construct short CoT data for simple problems and long CoT data for complex problems. The model begins with an initial ratio and performs reasoning compression using mixed data. After completing a compression cycle, the model re-evaluates the expected benefits of System-1 CoT data and System-2 CoT data to achieve improved performance. Specifically, and in line with intuition, System-1 CoT data can enhance efficiency, so we use an efficiency metric to measure the expected benefit of System-1 data. System-2 CoT, on the other hand, improves reasoning accuracy, and we use an accuracy metric to measure the benefit of System-2 data in terms of reasoning capability.

Compared to various methods requiring fine-tuning data with different reasoning lengths, our approach enables dynamic ratio learning by utilizing the self-sampled long CoT data and the short CoT data constructed by the original instruct/base model. Through experiments on DeepSeek-Distill-7B/14B, our model achieves excellent compression results on the 7B/14B models, with only a slight decrease in reasoning capability.
\section{Rethinking Short-Long CoT in Thinking Compression} 
We first constructed the short CoT data using simple problems and recorded how, as training steps increased, this subset contributed to token compression and accuracy retention across datasets of varying difficulty in math benchmarks.
\label{pre_section:takeaway1_2}

\begin{figure}
    \centering
    \includegraphics[width=1.0\linewidth]{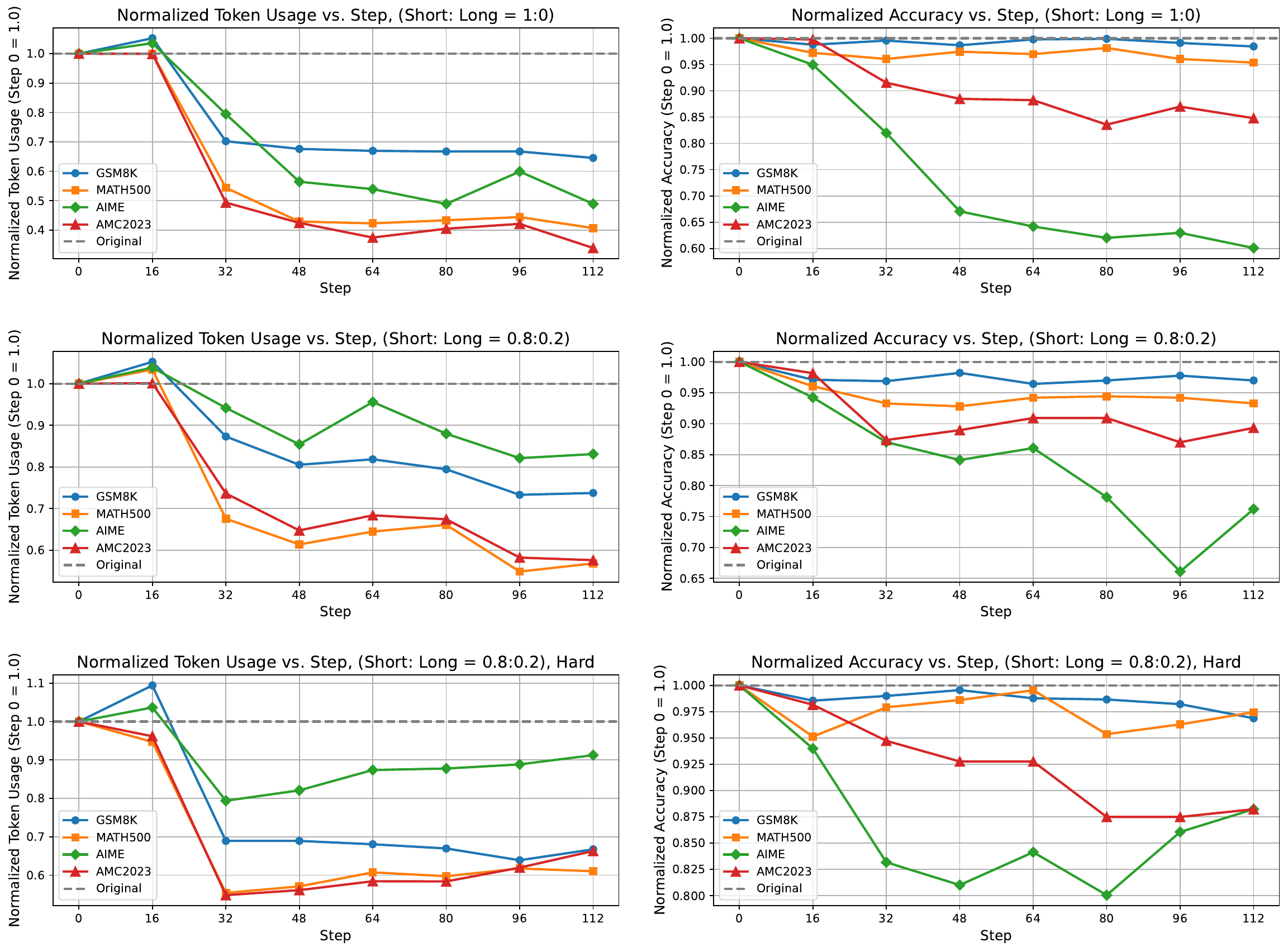}
    \caption{Impact of Combining Short CoT and Long CoT in Fixed Ratios on Thinking Compression Performance and Token Cost. We assessed the variation decay rate in output token length and accuracy on datasets of various question difficulty, spanning from GSM8K to AIME. The Normalized Token/Acc metric detail please refer to Equ. \ref{equ:normlized_acc} and Equ. \ref{equ:normlized_token}.}
    \label{fig:enter-label}
\end{figure}

\textbf{We find that short CoT thinking data for simple problems (System-1 data) can help compress the token usage across questions of various difficulty levels}. We leverage the short-cut solutions obtained from simple questions in GSM8K to fine-tune the model and then observe the token compression rates and accuracy drop rates across four datasets, ranging from simple to difficult: GSM8K, MATH500, AMC, and AIME. As shown in Figure \ref{fig:enter-label}, directly fine-tuning the long CoT model with short CoT data achieves good length compression for both simple and complex problems. We were pleasantly surprised to see that this form of length compression generalizes well across questions of all difficulty levels, and that it maintains strong performance on simple questions. However, this approach comes at a cost, as it leads to a significant decrease in reasoning ability on difficult problems. As this portion of the data is derived from intuitive CoT reasoning on simple problems, we denote it as System-1 data. It seems that directly using short CoT fine-tuning can only encourage the reasoning LLM to retain its System 1 reasoning abilities, while its ability for System 2 reasoning—slow and cautious thinking for complex problems—is largely lost.

\textbf{We find that long CoT thinking data for difficult problems (System-2 data) can help maintain the model's performance on challenging tasks, while simple question doesn't help much}. We sample with the s1~\cite{s1} like hard question prompt and then blend the System-2 data into the previous System-1 thinking dataset at a fixed short CoT \textit{vs.} long CoT ratio: $0.8$:$0.2$. We then observe the token compression rates and accuracy drop rates across four datasets.

It is worth noting that, by contrast, when we mix more long CoT data from simpler questions, the model still experiences a significant drop in performance on difficult questions. 
Refer to the middle and bottom parts of Figure \ref{fig:enter-label}, where we mix the long CoT sampled from challenging problems with the short CoT from simple problems. As a baseline, we also mix long CoT and short CoT from simple problems. The long CoT from difficult problems achieves lower accuracy drop rates across different datasets while maintaining comparable token compression rates.
We are unable to recover the original performance simply by using long CoT data from simple questions through data replay. Similar to the deliberate reasoning characteristic of the System-2 process on difficult problems, we refer to this part of the data as System-2 data.

A key question we directly address is \textbf{\textit{whether a direct mixing ratio of the two types of data(\textit{System-1/2} data), can be employed for post-training the long CoT model, resulting in a solution that eliminates redundancy}} without compromising performance. Based on these observations, we propose a dynamic approach aimed at identifying the optimal Thinking Compression data.

\begin{figure*}
    \centering
    \includegraphics[width=0.9\linewidth]{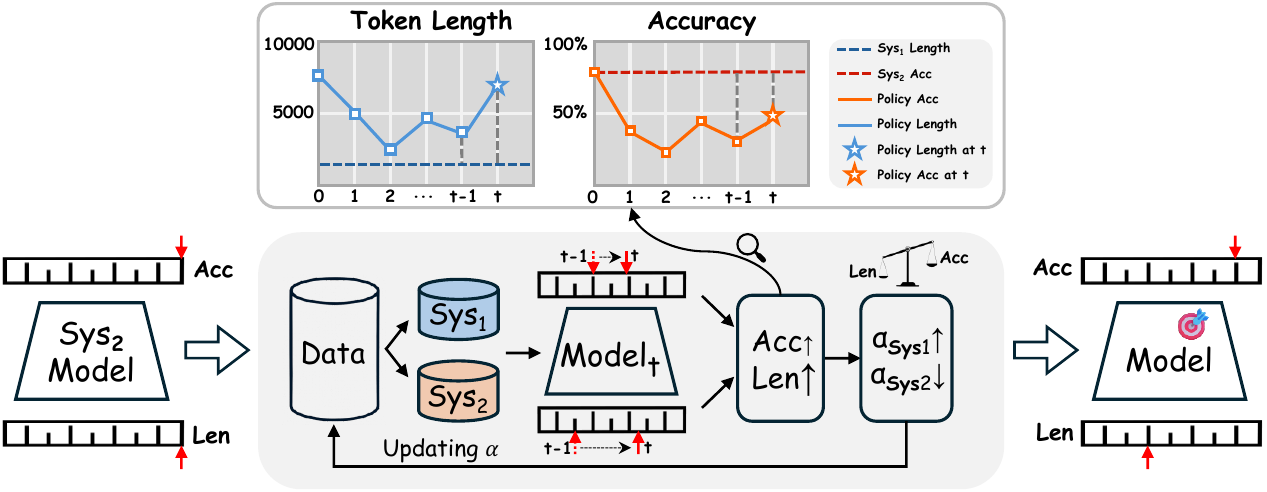}
    \caption{Overview of \name: Starting with a System-2 model, we iteratively update it on both Short-CoT and Long-CoT samples. The ratios of both data sources are adjusted every several steps based on the current average model accuracy and token length from the validation set until convergence.}
    \label{fig:123}
\end{figure*}

\section{Thinking Length Dynamic Re-weighting}
\vspace{-2mm}
\subsection{Short-Long CoT Reweighting with Relaxed Optimization}
\begin{algorithm*}[t]
\caption{Long-to-Short (L2S) Dynamic Reweighting Pipeline}
\label{alg:alg1}
\begin{algorithmic}
\Require Domain data $\mathcal{D}_{\text{system-1}}, \mathcal{D}_{\text{system-2}}, \mathcal{D}_{\text{dev}}$; training steps $T$; batch size $b$; step size $\eta$; smoothing parameter $c \in [0,1]$ (e.g., $c=10^{-4}$ in our implementation)
\State Initialize proxy weights $\theta_0$
\State Initialize mixture weights $\alpha_0 = (1/2, 1/2)$
\For{$t = 1$ to $T$}
    \State Let $|x|$ denote token length of example $x$ (with $|x| \leq L$)
    \State Compute benefit of fine-tuning with System-1 data:  $\lambda_{\text{system-1}}$ and System-2 data $\lambda_{\text{system-2}}$
    \State Update weights (entrywise exponential): $\alpha'_t \gets \alpha_{t-1} \cdot \exp(\eta \cdot \lambda_t)$
    \State Renormalize and smooth: $\alpha_t \gets (1-c)\frac{\alpha'_t}{\sum_{i=1}^k \alpha'_t[i]} + cu$
    \State Update proxy model weights $\theta_t$ using $L(\theta_{t-1}, \alpha_t)$ (e.g., via Adam, Adafactor)
\EndFor
\State \Return $\frac{1}{T} \sum_{t=1}^{T} \alpha_t$
\end{algorithmic}
\end{algorithm*}

\label{section:dynamic_takeaway}
We formalize the thinking compression problem as an optimization task to determine the optimal ratio between System-1 and System-2 reasoning.
We expect the model trained on mixed data to approach the superior performance of System-1 and System-2 in specific evaluation metrics. For model $M$ and input problem $x$, we define $T(y), C(y)$ as the token length and correctness of LLM output text $y$. 
We represent the System-1/2 ability bound as $\phi_{\text{sys-}i,\text{bound}}(x)$, in the following sections, we will abbreviate as $\phi_{\text{sys-}i,\text{bound}}(x)$
\begin{align}
\label{optimization_object}
&\min_{\theta,\, \alpha \in (0,1)}\; L(\theta, \alpha) = \sum_{i=1}^2 \alpha_i \cdot \delta_i \\
&\delta_i = \phi_{\text{sys-}i,\text{bound}}(x) - \phi_{\text{sys-}i,\theta}(x)
\end{align}
of which, $\phi_{sys-1, \theta}$ can be regarded as a metric for measuring the efficiency of the System-1 models.  $\phi_{sys-2, \theta}$ can be regarded as an accuracy metric.
In this way, the overall optimization objective is to minimize the gap between the model and the efficiency upper bound of System-1, as well as the reasoning capability upper bound of System-2, while simultaneously optimizing the model parameters to maximize both reasoning performance and efficiency. \\
\begin{align}
\phi_{sys-1, \text{bound}} &= -\mathbb{E}_{dev}[T(M_{s}(x))] \\
\phi_{sys-2, \text{bound}} &= \mathbb{E}_{dev}[C(M_{l}(x))]
\end{align}
\vspace{-5mm}

\paragraph{Setup for System-1/2 Mixed Data.}
Since System-1 can provide fast and intuitive answers to simple problems, we use the short CoT model to modulate the data for the System-1 model.
Since System-2 is designed to execute slow, logical reasoning for challenging problems, we employ the long CoT model to sample prompts from S1 ~\cite{s1}, retaining only the correct responses. Finally, we obtain $D_{system-1}=$<\textit{Simple Question}, \textit{Short CoT}> instruction pairs. For the harder problems within the System-1 domain, we used the long CoT model for sampling, resulting in a large amount of $D_{system-2}=$<\textit{Hard Question}, \textit{Long CoT}> instruction data. 

\subsection{Long-to-Short Data-Reweighting Tuning.}

\paragraph{Step 1: Estimate the ideal upper bounds of efficiency and performance.}
During training, we aim to continuously adjust the ratio of System-1 and System-2 data in the post-training phase, ensuring that the model retains the reasoning capabilities of the original long CoT model while achieving the efficiency of the short CoT model. 
Therefore, we set the accuracy upper bound, $\phi_{sys-2, \text{bound}}$, of the model obtained through mixed training to match the accuracy of the original long CoT model, while setting the token lower bound, $\phi_{sys-1, \text{bound}}$, of the mixed model to correspond to the data lower bound of the short CoT model we constructed.
\begin{align}
\phi_{\text{sys-2, bound}} &= \phi_{\text{sys-2}, L} = \hat{\mathbb{E}}_{\text{dev}}[C^{L}(x)] = \frac{1}{K} \sum_{i=1}^{K} \mathbbm{1}[\text{Correct}(y_i^{L})] \\ 
\phi_{\text{sys-1, bound}} &= \phi_{\text{sys-1}, short} = -\hat{\mathbb{E}}_{\text{dev}}[T^{S}(x)] = -\frac{1}{K} \sum_{i=1}^{K} \text{Token}(y_i^{S})
\end{align}

\vspace{-5mm}

\paragraph{Step 2: Thinking Compression Post-Train with dynamic System-1/2 reasoning weights}
We dynamically evaluate the utility of System-1 and System-2 reasoning data during training, and, guided by the performance of a reference model, adjust the sampling ratio between the two data types in real time to optimize training effectiveness.

\begin{align}
\lambda_{\text{sys-1}} &= \max\left( \frac{\phi_{\text{sys-1, bound}} - \phi_{\text{sys-1}, \theta_{\text{proxy}}}}{\phi_{\text{sys-1}, \theta_{s}} - \phi_{\text{sys-1}, \theta_{l}}},\ 0 \right)  \\
\lambda_{\text{sys-2}} &= \max\left( \frac{\phi_{\text{sys-2, bound}} - \phi_{\text{sys-2}, \theta_{\text{proxy}}}}{\phi_{\text{sys-2}, \theta_{l}} - \phi_{\text{sys-2}, \theta_{s}}},\ 0 \right)
\end{align}

\section{Experiments}
\label{sec:experiments}

\begin{table*}[t]
\centering
\small
\resizebox{\textwidth}{!}{%
  \setlength\tabcolsep{2pt}
  \begin{tabular}{lcccccccccccccc}
  \toprule
  \multirow{2}{*}{\textbf{Model}} & \multicolumn{7}{c}{\textbf{Accuracy}} & \multicolumn{6}{c}{\textbf{Generation Length}} & \multirow{2}{*}{\textbf{A.C.R.}} \\
  \cmidrule(r){2-8} \cmidrule(r){9-14}
  & \makecell{ASDiv} & \makecell{GSM8K}  & \makecell{MATH} & \makecell{AIME} & \makecell{AMC} & \makecell{Minerva} & \cellcolor{gray!20}\makecell{Avg.}
  & \makecell{ASDiv} & \makecell{GSM8K}  & \makecell{MATH} & \makecell{AIME} & \makecell{AMC} & \makecell{Minerva}
  & \\
  \midrule \midrule
  \multicolumn{15}{c}{\textbf{\textit{7B Models}}} \\
  \midrule
  R1-Distill-Qwen & 86.8 & 89.4  & 86.8 & 42.9 & 81.5 & 46.0 & \cellcolor{gray!20}72.2 & 769 & 554  & 2861 & 6820 & 4510 & 3347 & -- \\
  \rowcolor{yellow!20}
  TALE-EP & 80.4 & 89.1  & 84.3 & 40.0 & 80.0 & 42.3 & \cellcolor{gray!20}69.3 & 509  & 450 & 1994 & 6520 & 3892 & 2242 & 22.3\% \\
  \rowcolor{yellow!20}
  ConciseCoT & 86.0 & 89.5  & 86.2 & 41.7 & 79.6 & 46.0 & \cellcolor{gray!20}71.5 & 532 & 457  & 2330 & 6587 & 4245 & 3347 & 12.7\% \\
  \rowcolor{green!20}
  Avg. Merging & 92.8 & 70.1  & 58.6 & 0.05 & 39.6 & 29.8 & \cellcolor{gray!20}48.4 & 622 & 8552  & 8540 & 8501 & 8542 & 8544 & 3.2\% \\
  \rowcolor{green!20}
  Task-Arithmetic-Merging & 83.3 & 84.6  & 74.6 & 20.0 & 63.5 & 39.6 & \cellcolor{gray!20}61.0 & 321 & 383  & 907 & 2500 & 1311 & 794 & 61.3\% \\
  \rowcolor{green!20}
  Ties-Merging & 74.4 & 69.7  & 59.8 & 13.6 & 42.5 & 23.2 & \cellcolor{gray!20}47.2 & 1114 & 2475  & 4086 & 6767 & 5195 & 4306 & 0.1\% \\
  \rowcolor{green!20}
  Ties-Dare-Merging & 75.9 & 72.3  & 65.4 & 14.6 & 45.6 & 24.3 & \cellcolor{gray!20}49.6 & 1036 & 2073  & 2934 & 5483 & 3698 & 2938 & 8.3\% \\
  \rowcolor{red!10}
  Overthink & 86.6 & 89.6  & 87.2 & 38.7 & 79.6 & 45.2 & \cellcolor{gray!20}71.1 & 773 & 555  & 2898 & 6766 & 4558 & 3407 & 0.1\% \\
  \rowcolor{red!10}
  ThinkPrune & 90.6 & 92.1  & 91.0 & 43.3 & 86.2 & 45.6 & \cellcolor{gray!20}74.8 & 653 & 587  & 2379 & 6207 & 3739 & 2762 & 12.6\% \\
  \rowcolor{red!10}
  $\text{CoT-Valve}^{*}$ & 59.4 & 88.4  & 84.2 & 41.2 & 80.6 & 41.9 & \cellcolor{gray!20}65.9 & 140 & 514  & 2144 & 6397 & 4278 & 2172 & 26.8\% \\
  \hdashline
  \name & 93.0 & 87.7  & 87.4 & 41.2 & 83.1 & 44.5 & \cellcolor{gray!20}72.8 & 147 & 253  & 1556 & 6368 & 3386 & 1451 & 44.9\% \\
  \rowcolor{blue!10}
  $\Delta$ & +6.2 & -1.7  & +0.6 & -1.7 & +1.6 & -1.5 & \cellcolor{gray!20}+0.7 & -622 & -301  & -1305 & -452 & -1124 & -1896 & -- \\
  \midrule \midrule
  \multicolumn{15}{c}{\textbf{\textit{14B Models}}} \\
  \midrule
  R1-Distill-Qwen & 80.5 & 92.5  & 86.4 & 43.4 & 79.6 & 48.2 & \cellcolor{gray!20}71.7 & 476 & 679  & 2951 & 6701 & 4584 & 3270 & -- \\
  \rowcolor{yellow!20}
  TALE-EP & 77.5 & 92.4  & 85.4 & 49.2 & 80.3 & 50.0 & \cellcolor{gray!20}72.5 & 369 & 555  & 2248 & 6551 & 4179 & 2731 & 15.4\% \\
  \rowcolor{yellow!20}
  ConciseCoT & 74.0 & 92.4  & 85.6 & 51.6 & 82.3 & 47.1 & \cellcolor{gray!20}72.2 & 369 & 555  & 2066 & 6267 & 3878 & 2605 & 18.8\% \\
  \rowcolor{green!20}
  Avg. Merging & 94.8 & 90.3  & 73.0 & 10.8 & 55.0 & 44.1 & \cellcolor{gray!20}61.3 & 167 & 366  & 5158 & 6364 & 5668 & 1084 & 30.5\% \\
  \rowcolor{green!20}
  Task-Arithmetic-Merging & 86.5 & 86.5  & 74.2 & 13.3 & 55.3 & 36.0 & \cellcolor{gray!20}58.6 & 238 & 368  & 870 & 2813 & 1411 & 1050 & 60.2\% \\
  \rowcolor{green!20}
  Ties-Merging & 79.6 & 91.3  & 82.6 & 25.4 & 72.5 & 37.1 & \cellcolor{gray!20}64.8 & 242 & 542  & 1919 & 5913 & 3158 & 1850 & 31.8\% \\
  \rowcolor{green!20}
  Ties-Dare-Merging & 80.7 & 91.8  & 84.8 & 25.4 & 75.3 & 34.9 & \cellcolor{gray!20}65.4 & 274 & 467  & 1870 & 5747 & 3182 & 1877 & 33.0\% \\
  \rowcolor{red!10}
  Overthink & 79.3 & 92.3  & 88.0 & 45.8 & 82.8 & 45.6 & \cellcolor{gray!20}72.3 & 451 & 679  & 2893 & 6700 & 4464 & 3715 & 1.6\% \\
  \rowcolor{red!10}
  ThinkPrune & 80.6 & 93.7  & 89.0 & 50.8 & 88.7 & 50.7 & \cellcolor{gray!20}75.6 & 379 & 563  & 2177 & 5778 & 3327 & 2234 & 22.8\% \\
  \rowcolor{red!10}
  $\text{CoT-Valve}^{\&}$ & 72.9 & 92.0  & 87.0 & 45.0 & 83.5 & 47.8 & \cellcolor{gray!20}71.4 & 204 & 576  & 2652 & 6686 & 4392 & 2833 & 16.7\% \\
  \hdashline
  \name & 88.0 & 90.9  & 86.6 & 43.3 & 83.8 & 48.7 & \cellcolor{gray!20}73.5 & 158 & 240  & 2092 & 6403 & 3839 & 2177 & 35.8\% \\
  \rowcolor{blue!10}
  $\Delta$ & +8.0 & -1.6  & +0.2 & -0.1 & +4.2 & +0.5 & \cellcolor{gray!20}+2.1 & -318 & -439  & -859 & -298 & -745 & -1093 & -- \\
  \bottomrule
  \end{tabular}%
}
\caption{
Performance comparison of \name with baselines. The accuracy is measured by sampling multiple responses from the LLMs and taking the average to reduce variance. * denotes the CoT-Valve~\cite{CoT-Valve} result that we reproduced using the official dataset. $\Delta$ refers to \name in comparison with \textit{Original}. Math' and Minerva' refer to MATH500 and MinervaMath datasets, respectively. A.C.R. means the token compression ratio computed by Eq.~\ref{equ:compreesion_rate}. In the table: \textcolor{yellow!60!white}{yellow} represents \textit{prompt-based} methods; \textcolor{green!60!white}{green} highlights Merging-based methods; 
\textcolor{red!60!white}{red} indicates \textit{SFT-based} and \textit{RL-based} methods.
}
\label{tab:main_exp}
\end{table*}
\subsection{Experimental Setup}
\label{sec:setup}

\begin{table*}[t]
\centering
\renewcommand{\arraystretch}{1.0}
\resizebox{\textwidth}{!}{
\begin{tabular}{l|cccccccccccc}
\toprule[1.5pt]
\multirow{2}{*}{\textbf{Model}} & \multicolumn{6}{c}{\textbf{Accuracy}} & \multicolumn{6}{c}{\textbf{Generation Length}} \\
\cmidrule(lr){2-7} \cmidrule(lr){8-13}
& \makecell{GSM8K} & \makecell{MATH} & \makecell{AIME} & \makecell{AMC} & \makecell{Minerva}  & \cellcolor{gray!20}Avg.  
& \makecell{GSM8K} & \makecell{MATH} & \makecell{AIME} & \makecell{AMC} & \makecell{Minerva} & \cellcolor{gray!20}Avg. \\
\midrule
\multicolumn{13}{c}{\textbf{\textit{7B Models}}} \\
\midrule
Original Model   & 89.4 & 86.8 & 42.9 & 81.5 & 46.0  & \cellcolor{gray!20}69.3 & 554 & 2861 & 6820 & 4510 & 3347 & \cellcolor{gray!20}3618 \\
\hdashline
-\textit{MixChain-Z-GSM8K}\textsuperscript{\&} & 88.4 & 84.2 & 41.2 & 80.6 & 41.9  & \cellcolor{gray!20}67.3 & 514 & 2144 & 6397 & 4278 & 2172  & \cellcolor{gray!20}3101  \\
-\textit{Static-Mixture} & 87.1  & 84.8  & 39.7 & 73.1  & 35.5 & \cellcolor{gray!20}64.0 & 236 & 1221 & 5322 & 2560 & 1544 & \cellcolor{gray!20}2177  \\
-\textit{\name}  & 87.7 & 87.4 & 41.2 & 83.1 & 41.0 & \cellcolor{gray!20}68.1 & 253 & 1556 & 6368 & 3386 & 1434 & \cellcolor{gray!20}2599 \\
\bottomrule[1.5pt]
\end{tabular}
}
\caption{Performance comparison of \name with static baselines. The accuracy is measured by sampling multiple responses from the LLMs and taking the average to reduce variance. \textsuperscript{\&} denotes the CoT-Valve~\cite{CoT-Valve} result that we reproduced using the official dataset. Math and Minerva mean MATH500 and MinervaMath datasets.}
\label{tab:dynamic_aba}
\end{table*}

\begin{table*}[t]
\centering
\renewcommand{\arraystretch}{1.0}
\resizebox{\textwidth}{!}{
\begin{tabular}{lcccccccccc}
\toprule[1.5pt]
\multirow{2}{*}{\textbf{Model}} & \multicolumn{5}{c}{\textbf{Accuracy}} & \multicolumn{5}{c}{\textbf{Generation Length}} \\
\cmidrule(r){2-6} \cmidrule(r){7-11}
& \makecell{GSM8K} & \makecell{MATH} & \makecell{AIME} & \makecell{AMC}  & \cellcolor{gray!20}Avg. & \makecell{GSM8K} & \makecell{MATH} & \makecell{AIME} & \makecell{AMC} & \cellcolor{gray!20}Avg. \\
\midrule
\multicolumn{11}{c}{\textbf{\textit{DeepSeek-R1-Distill-Qwen-7B}}} \\
\midrule
Original Model   & 89.4 & 86.8 & 42.9 & 81.5  & \cellcolor{gray!20}75.2 & 554 & 2861 & 6820 & 4510 & \cellcolor{gray!20}3686 \\
\hdashline
-\textit{\name}     & 87.7 & 87.4 & 41.2 & 83.1 & \cellcolor{gray!20}74.8 & 253 & 1556 & 6368 & 3386 & \cellcolor{gray!20}2891 \\
-\textit{L1-same}   & 86.4 & 88.6 & 42.2 & 84.6 & \cellcolor{gray!20}75.4 & 301 & 2301 & 5875 & 3784 & \cellcolor{gray!20}3056 \\
-\textit{L1-lower}  & 86.4 & 87.6 & 45.1 & 84.6 & \cellcolor{gray!20}75.9 & 312 & 1831 & 5675 & 3807 & \cellcolor{gray!20}2906 \\
-\textit{L1-higher} & 86.1 & 88.4 & 45.5 & 83.3 & \cellcolor{gray!20}75.8 & 292 & 2589 & 6007 & 3746 & \cellcolor{gray!20}3158 \\
\bottomrule[1.5pt]
\end{tabular}
}
\caption{Performance comparison of \name with budget-aware baseline, L1~\citep{L1}. 
The accuracy is measured by sampling multiple responses from the LLMs and to reduce variance. The terms \textit{same}, \textit{lower}, and \textit{higher} refer to setting the budget to match our results, 20\% lower, and 20\% higher, respectively. MATH means MATH500 dataset.}
\label{tab:bedget_model_compare}
\end{table*}

\begin{table*}[t]
\centering
\renewcommand{\arraystretch}{1.0}
\resizebox{\textwidth}{!}{
\begin{tabular}{lcccccccccc}
\toprule[1.5pt]
\multirow{2}{*}{\textbf{Model}} & \multicolumn{5}{c}{\textbf{Accuracy}} & \multicolumn{5}{c}{\textbf{Generation Length}} \\
\cmidrule(r){2-6} \cmidrule(r){7-11} 
& \makecell{GSM8K} & \makecell{MATH} & \makecell{AIME} & \makecell{AMC} & \cellcolor{gray!20}Avg.  
& \makecell{GSM8K} & \makecell{MATH} & \makecell{AIME} & \makecell{AMC} & \cellcolor{gray!20}Avg. \\
\midrule
\multicolumn{11}{c}{\textbf{\textit{DeepSeek-R1-Distill-Qwen-7B System-1 Short CoT Ablation}}} \\
\midrule
Original Model   & 89.4 & 86.8 & 42.9 & 81.5 & \cellcolor{gray!20}60.1 & 554 & 2861 & 6820 & 4510 & \cellcolor{gray!20}2949 \\
-\textit{\name-Easy}   & 87.7 & 87.4 & 41.2 & 83.1 & \cellcolor{gray!20}59.9 & 253 & 1556 & 6368 & 3386 & \cellcolor{gray!20}2313 \\ 
-\textit{\name-Medium} & 88.2 & 86.2 & 41.5 & 31.3 & \cellcolor{gray!20}61.8 & 318 & 2083 & 6604 & 3945 & \cellcolor{gray!20}3238 \\
-\textit{\name-Hard}   & 83.6 & 80.2 & 30.0 & 65.3 & \cellcolor{gray!20}64.8 & 495 & 2970 & 6874 & 4947 & \cellcolor{gray!20}3822 \\
\midrule
\multicolumn{11}{c}{\textbf{\textit{DeepSeek-R1-Distill-Qwen-7B System-2 Long CoT Ablation}}} \\
\midrule
Original Model   & 89.4 & 86.8 & 42.9 & 81.5 & \cellcolor{gray!20}60.1 & 554 & 2861 & 6820 & 4510 & \cellcolor{gray!20}2949 \\
-\textit{\name-Easy}   & 83.9 & 86.8 & 42.5 & 83.4 & \cellcolor{gray!20}74.2 & 446 & 2639 & 6580 & 4047 & \cellcolor{gray!20}3428 \\
-\textit{\name-Medium} & 91.6 & 87.6 & 40.4 & 81.5 & \cellcolor{gray!20}75.3 & 542 & 2761 & 6553 & 4116 & \cellcolor{gray!20}2950 \\
-\textit{\name-Hard}   & 87.7 & 87.4 & 41.2 & 83.1 & \cellcolor{gray!20}74.8 & 253 & 1556 & 6368 & 3386 & \cellcolor{gray!20}2828 \\ 
\bottomrule[1.5pt]
\end{tabular}
}
\caption{An ablation study on the difficulty levels of ShortCoT and LongCoT was conducted during the construction of the ShortCoT and LongCoT dataset. The accuracy is measured by sampling multiple responses from the LLMs and taking the average to reduce variance. $^\&$ denotes the CoT-Valve \cite{CoT-Valve} result that we reproduced using the officially dataset. MATH means MATH500 dataset.}
\label{tab:aba_data_source}
\end{table*}

\begin{figure*}[t]
  \centering
  \begin{subfigure}[t]{0.24\linewidth}
    \centering
    \includegraphics[width=\linewidth]{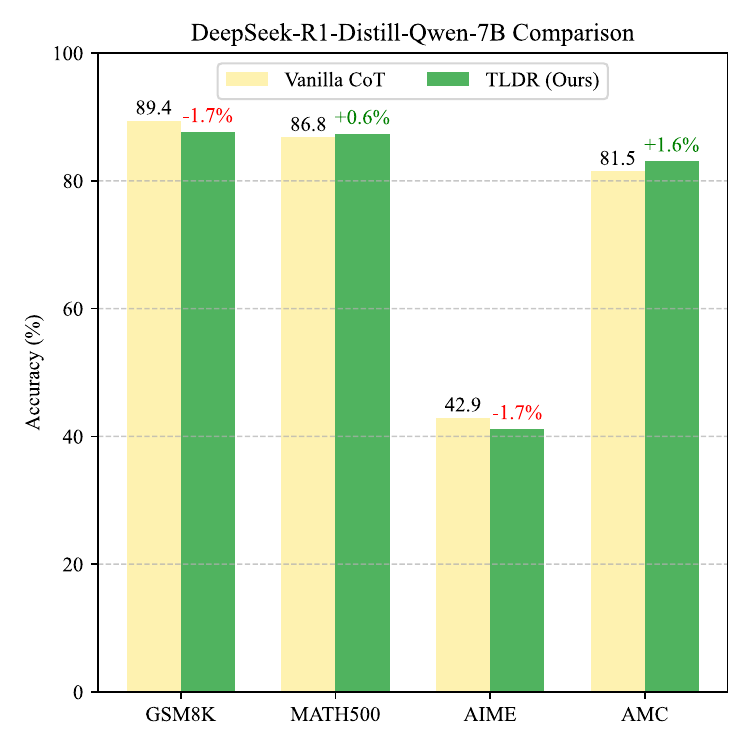}
    \vspace{-7mm}
  \end{subfigure}
  \begin{subfigure}[t]{0.24\linewidth}
    \centering
    \includegraphics[width=\linewidth]{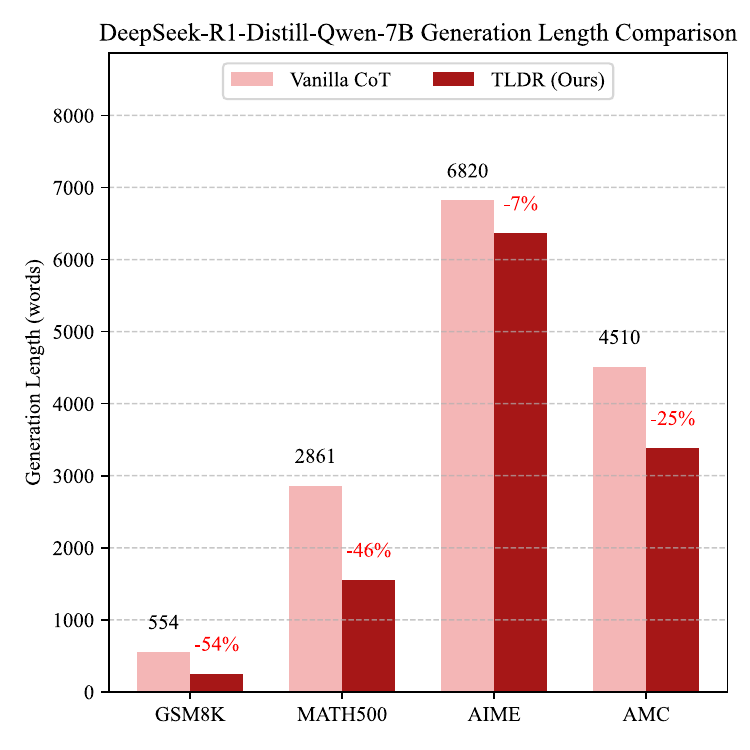}
    \vspace{-7mm}
  \end{subfigure}
  \begin{subfigure}[t]{0.24\linewidth}
    \centering
    \includegraphics[width=\linewidth]{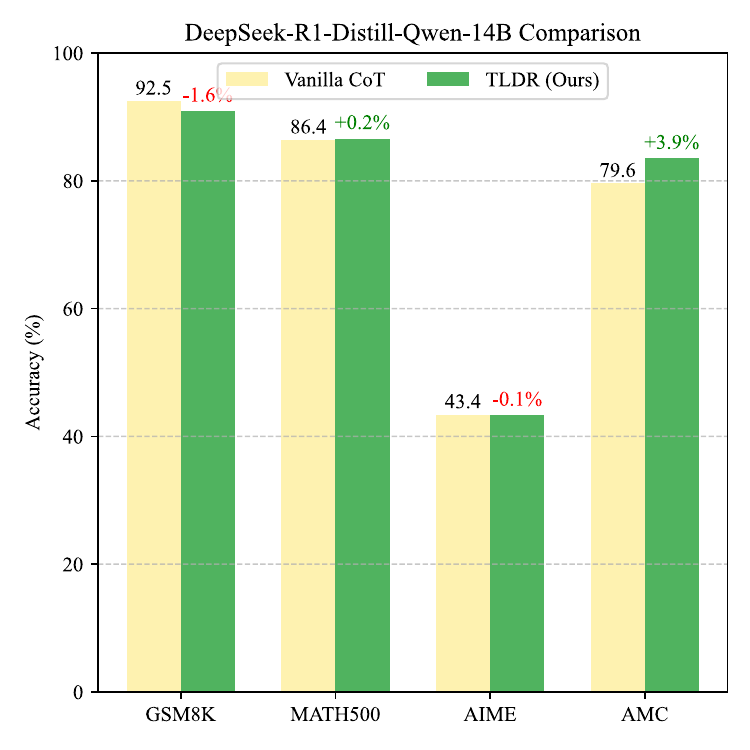}
    \vspace{-7mm}
  \end{subfigure}
  \begin{subfigure}[t]{0.24\linewidth}
    \centering
    \includegraphics[width=\linewidth]{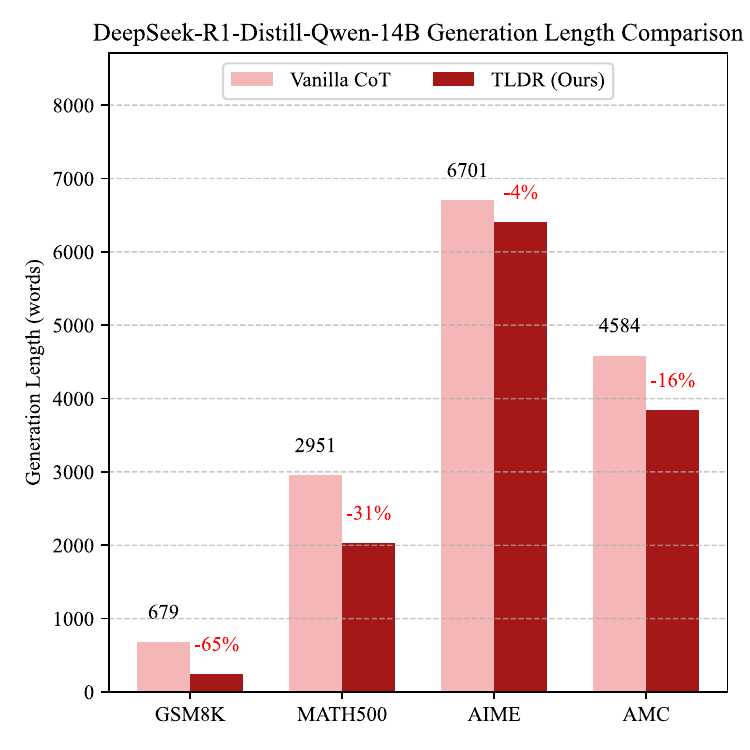}
    \vspace{-7mm}
  \end{subfigure}
  \caption{Comparison of accuracy and generation length between Vanilla CoT and our \name method on four benchmark datasets (GSM8K, MATH500, AIME, AMC) using DeepSeek-R1-Distill-Qwen models. \name consistently reduces generation length while maintaining or improving accuracy across both 7B and 14B model scales.}
  \label{fig:comparison}
  \vspace{-4mm}
\end{figure*}

\begin{figure*}
    \centering
    \includegraphics[width=1.0\linewidth]{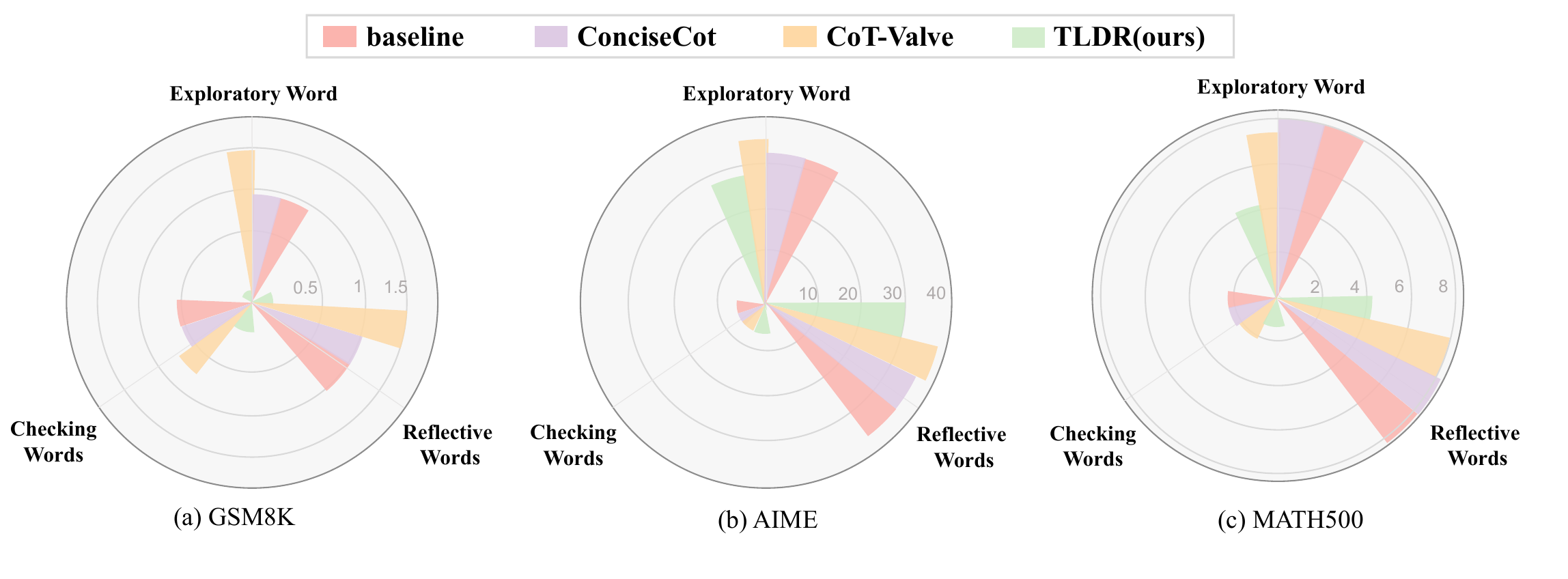}
    \vspace{-10mm}
    \caption{Frequency comparison of different keywords. The figure illustrates the distribution of exploratory, checking, and reflective keywords across datasets. \textit{Exploratory Keywords: wait}, \textit{Reflective Word: but}, \textit{Checking Words: make sure/confirm/verify/check}, \name significantly reduces the presence of such words, reflecting its ability to produce streamlined and efficient reasoning steps.}
    \label{fig:lidar}
    \vspace{-5mm}
\end{figure*}

\paragraph{Datasets and Metrics.}
Following prior efforts, we evaluate \name on several widely-used benchmarks that span a broad range of difficulty levels, including ASDiv~\cite{miao2021diverse}, GSM8K~\citep{cobbe2021training}, MATH-500~\citep{hendrycks2021measuring}, and AIME2024~\citep{AIME2024}, AMC~\citep{AMC2023} in Table~\ref{tab:main_exp}. To ensure the stability of the evaluation, we performed multiple samplings for each dataset and took the average accuracy. For GSM8K, MATH500, and GPQA, we sampled each question 4 times and took the average accuracy of the 4 solutions. For AIME24 and AMC23, we sampled each solution 8 times and took the average accuracy of the 8 solutions. The token count was calculated using the corresponding tokenizer of the language model, and our evaluation environment was modified using the Skythought library.

\paragraph{Baselines.}
We compared three types of baselines: 

\noindent \textit{Prompt-based} methods. We compared our approach with the well-known prompt-based baselines in the community, including TALE-EP~\cite{TALE-EP}, which requires the prompt to be as simple as possible, and ConciseCoT~\cite{ConciseCoT}, which demands the use of the most concise CoT steps during step-by-step reasoning.

\noindent \textit{Model-Merging-based Methods.} Model Merging leverages the rich knowledge from short CoT Instruct and the long CoT model for model fusion, aiming to achieve the shortest yet most effective reasoning process. We compared this approach with the Avg. Merging method used in Kimi-1.5~\cite{team2025kimi, Merging-Long2Short} and some advanced merging method, like Task-Arithmetic-Merging, Ties-Merging, Ties-Dare-Merging, discussed in~\cite{Merging-Long2Short}. 

\noindent \textit{Reward-based Methods.} ThinkPruner~\cite{thinkerprune} uses progressive compression of RL training length to improve the effectiveness of context utilization during exploration. $SimPO_{shortest}$ was introduced in Overthink~\cite{kumar2025overthink} to adjust the effectiveness of the RL algorithm by length-guided RL training.

We find that our approach can maintain accuracy across datasets of varying difficulty while achieving satisfactory compression rates. In contrast, merging-based methods still suffer from significant performance drops on challenging problems. Compared to CoT-Valve and ThinkPrune, our method attains excellent compression rates, particularly on the ASDiv and GSM8K datasets, where over-exploration tends to occur. CoT-Valve, as an SFT-based approach, requires carefully designed model blending and the construction of length-diverse datasets for dynamic learning. In contrast, our method only requires straightforward data sampling and adaptive mixing ratios, achieving adaptive reasoning in a much simpler way.

\vspace{-3mm}
\subsection{Comparison of Dynamic Compression vs. Static Compression Data.}
We compared our dynamic algorithm with several simple static data mixing methods, including directly mixing data according to a simple ratio and constructing length-uniform data using parameter interpolation. MixChain-Z-GSM8K is a Long2Short dataset proposed by CoT-Valve in Table \ref{tab:dynamic_aba}.

\subsection{Ablation of Different System-1/2 Source}

We also discovered that incorporating higher-difficulty CoT data into a short-long mixed dataset could effectively eliminate redundancies in CoT for their compressed version. However, direct mixing could lead to performance degradation. After introducing a dynamic ratio method, we found that flexibly adjusting the ratio could effectively maintain performance in Table \ref{tab:aba_data_source}. We categorized the sources of questions in the thinking compression data into three difficulty levels: \textit{easy}, \textit{medium}, and \textit{hard}. \textit{easy} questions are from GSM8K, \textit{medium} questions are from the training set of MATH500, and \textit{hard} questions are from the s1 prompt questions.

\noindent \textbf{Short CoT Compression Generalization Analysis of Easy-to-Hard.}
We tested the construction of System-1 data, examining the composition of data from different thinking compression sources. Our experiments found that constructing data based on low-difficulty problems could significantly reduce the token count of high-difficulty problems while maintaining performance. We found that using lower-difficulty problems to construct thinking compression data for redundancy removal can further generalize to higher-difficulty problems.

\noindent \textbf{Long CoT Performance Generalization Analysis of Hard-to-Easy}
We also conducted an analysis of the following aspects: during the sampling of long CoT, we utilized data from three distinct sources—\textit{easy}, \textit{medium}, \textit{hard} prompt. Our findings reveal that only by constructing long CoT using hard problems and dynamically adjusting their proportions during training can we recover the original performance associated with long CoT. This strategy effectively mitigates the risk of forgetting in reasoning capabilities during continual learning.


\subsection{Comparison with Token Budgeted-Aware Model}

We compared our redundancy reduction method with both quota-controlled models and reasoning models under the same token budget, in order to evaluate the effectiveness of our approach relative to explicit quota-based control. The results show that our method achieves higher reasoning accuracy than both the L1 \cite{L1} baselines under the same token quota. Furthermore, our approach demonstrates more efficient utilization of context length and does not require explicitly specifying a reasoning quota, offering a more flexible and adaptive inference mechanism. \name demonstrates stronger compression efficiency on simple problems.

\subsection{Analysis of Thinking Patterns: Reflections \& Solutions}
We compared our method with other thinking compression methods in terms of their impact on changes in cognitive patterns ~\cite{RedStar} of the solution in Figure \ref{fig:comparison}. We performed fine-grained statistical analysis on the results across different datasets and various levels of MATH500. Our analysis demonstrates that our approach effectively compresses the internal redundancy and reflects the properties of the solution patterns. \name effectively reduces the reliance on such macro reasoning patterns in benchmarks like GSM8K and MATH500, thereby avoiding excessive allocation of computational budget. Notably, for more challenging problems, the model still retains a significant degree of macro reasoning behavior to preserve its System-2 reasoning capabilities.

\vspace{-2mm}

\section{Related Work}

\paragraph{Efficient System-2 Reasoning.}
Despite the strong generalization and reasoning abilities demonstrated by the system 2 reasoning paradigm, the auto-regressive nature of LLMs imposes a significant reasoning burden, limiting their application in agent-based or edge scenarios~\cite{OS-ATLAS}. To address this, various approaches have emerged to improve the reasoning efficiency. These methods can be broadly categorized into two types. One category focuses on building \textit{adaptive reasoning-budget}. Within this, some training-free methods like CoD~\cite{CoD} and TALE-EP~\cite{TALE-EP} impose budget constraints to control overall reasoning cost. Budget-sensitive models such as L1~\cite{L1}, TOPS~\cite{TOPS}, o1 Pruner~\cite{o1_pruner}, and K1.5~\cite{team2025kimi} add length penalties during the post-training. Some work~\cite{CoT-Valve, jiang2025drp,wang2024explore,chain-of-reasoning} synthesizes diverse-length CoT data, while TOPS~\cite{TOPS} samples budget-sensitive versions using a data model, and C3oT~\cite{C3oT} compresses original LLM output to train jointly on shorter CoTs. Other approaches involve external models or switching mechanisms~\cite{Reasoning-in-Flux} to allocate budgets. Routellm~\cite{RouteLLM} uses multiple routers to find the most suitable reasoning model, while Self-REF~\cite{Self-REF} employs confidence scores to route based on reasoning difficulty. 
Another category concentrates on building \textit{efficient representations}. TokenSkip~\cite{TokenSkip} selects data based on token importance for compressed reasoning and more concise thought chains. COCONUT~\cite{COCONUT} explores more efficient reasoning in the latent space. ICoT-KD~\cite{CCoT} and CCoT~\cite{CCoT} attempt to build more efficient reasoning strategies in the hidden space, while Token Assorted combines hidden space and text-based reasoning to balance interpretability and efficiency. Heima~\cite{Heima} extends hidden-space reasoning to multi-modal models. Unlike prior efforts, \name compresses the length of the reasoning chain without introducing complex data construction and without compromising the original reasoning LLM's inferential representations.

\paragraph{Data Balancing in Pre/Post Training.}

The quality and proportion of data are critical during both the pre-training and post-training phases. In the pre-training stage, data quality and proportion are primarily managed through filtering and reweighting. Pre-training data filtering, extensively studied to boost model performance and training efficiency~\cite{liu2024datasets, albalak2024survey}, typically involves steps like language filtering~\cite{bigscience, chowdhery2022palm}, quality filtering~\cite{raffel2023exploring, rae2022scaling}, content filtering~\cite{xu2021detoxifying, longpre2023pretrainers}, and deduplication~\cite{hernandez2022scaling, lee2022deduplicating}. While these methods significantly enhance corpus quality, their static nature can hinder dynamic adjustments during training, potentially discarding valuable data~\cite{muennighoff2023scaling} and introducing biases~\cite{gururangan-etal-2022-whose, longpre2023pretrainers, dodge2021documenting}. Similarly, in the post-training stage, an appropriate proportion of data with varying characteristics is crucial for optimizing final performance~\cite{bpo,zhang_balancing}. For example, DeepMath-103K generates a large volume of data with evenly distributed difficulty for training~\cite{DeepMath-103K}, SRPO designs a dynamic sampling approach to filter out samples that are consistently answered correctly, thereby improving inference efficiency~\cite{SPRO}. EffiCode~\citep{huang2024effi} develop a self-optimization process based on overhead profiling and iterative refinement, resulting in a high-quality dataset for fine-tuning LLMs to produce more efficient and accurate code. To the best of our knowledge, we are the first to introduce a re-weighting mechanism into thinking compression. By employing simple strategies to construct short and long CoT, we enable the model to dynamically compress its reasoning process.

\section{Conclusion}

This paper introduces \name, an innovative method designed to compress the reasoning processes of LLMs without sacrificing accuracy. By dynamically re-weighting the influence of system 1 (concise reasoning) and system 2 (detailed reasoning) data during the training process, \name allows LLMs to eliminate unnecessary steps for simpler problems while still engaging in deep contemplation for complex tasks. \name avoids the laborious data collection and hyperparameter tuning typically required by other compression methods, offering a more practical solution for developing LLMs that are both efficient and accurate.
\clearpage
\bibliographystyle{plain}
\bibliography{neurips}

\begin{thebibliography}{10}

\bibitem{L1}
Pranjal Aggarwal and Sean Welleck.
\newblock L1: Controlling how long a reasoning model thinks with reinforcement learning.
\newblock {\em arXiv preprint arXiv:2503.04697}, 2025.

\bibitem{AIME2024}
AI-MO.
\newblock {Aime 2024}, 2024.

\bibitem{AMC2023}
AI-MO.
\newblock {Amc 2023}, 2024.

\bibitem{albalak2024survey}
Alon Albalak, Yanai Elazar, Sang~Michael Xie, Shayne Longpre, Nathan Lambert, Xinyi Wang, Niklas Muennighoff, Bairu Hou, Liangming Pan, Haewon Jeong, Colin Raffel, Shiyu Chang, Tatsunori Hashimoto, and William~Yang Wang.
\newblock A survey on data selection for language models, 2024.

\bibitem{Overthink}
Xingyu Chen, Jiahao Xu, Tian Liang, Zhiwei He, Jianhui Pang, Dian Yu, Linfeng Song, Qiuzhi Liu, Mengfei Zhou, Zhuosheng Zhang, Rui Wang, Zhaopeng Tu, Haitao Mi, and Dong Yu.
\newblock Do not think that much for 2+3=? on the overthinking of o1-like llms, 2025.

\bibitem{CCoT}
Jeffrey Cheng and Benjamin Van~Durme.
\newblock Compressed chain of thought: Efficient reasoning through dense representations.
\newblock {\em arXiv preprint arXiv:2412.13171}, 2024.

\bibitem{chowdhery2022palm}
Aakanksha Chowdhery, Sharan Narang, Jacob Devlin, Maarten Bosma, Gaurav Mishra, Adam Roberts, Paul Barham, Hyung~Won Chung, Charles Sutton, Sebastian Gehrmann, et~al.
\newblock Palm: Scaling language modeling with pathways, 2022.

\bibitem{Self-REF}
Yu-Neng Chuang, Helen Zhou, Prathusha Sarma, Parikshit Gopalan, John Boccio, Sara Bolouki, and Xia Hu.
\newblock Learning to route llms with confidence tokens.
\newblock {\em arXiv preprint arXiv:2410.13284}, 2025.

\bibitem{cobbe2021training}
Karl Cobbe, Vineet Kosaraju, Mohammad Bavarian, Mark Chen, Heewoo Jun, Lukasz Kaiser, Matthias Plappert, Jerry Tworek, Jacob Hilton, Reiichiro Nakano, et~al.
\newblock {Training verifiers to solve math word problems}.
\newblock {\em arXiv preprint arXiv:2110.14168}, 2021.

\bibitem{Deepseek-R1}
DeepSeek-AI, Daya Guo, Dejian Yang, Haowei Zhang, Junxiao Song, Ruoyu Zhang, Runxin Xu, Qihao Zhu, Shirong Ma, Peiyi Wang, Xiao Bi, Xiaokang Zhang, Xingkai Yu, Yu~Wu, Z.~F. Wu, Zhibin Gou, Zhihong Shao, Zhuoshu Li, Ziyi Gao, Aixin Liu, Bing Xue, Bingxuan Wang, Bochao Wu, Bei Feng, Chengda Lu, Chenggang Zhao, Chengqi Deng, Chenyu Zhang, Chong Ruan, Damai Dai, Deli Chen, Dongjie Ji, Erhang Li, Fangyun Lin, Fucong Dai, Fuli Luo, Guangbo Hao, Guanting Chen, Guowei Li, H.~Zhang, Han Bao, Hanwei Xu, Haocheng Wang, Honghui Ding, Huajian Xin, Huazuo Gao, Hui Qu, Hui Li, Jianzhong Guo, Jiashi Li, Jiawei Wang, Jingchang Chen, Jingyang Yuan, Junjie Qiu, Junlong Li, J.~L. Cai, Jiaqi Ni, Jian Liang, Jin Chen, Kai Dong, Kai Hu, Kaige Gao, Kang Guan, Kexin Huang, Kuai Yu, Lean Wang, Lecong Zhang, Liang Zhao, Litong Wang, Liyue Zhang, Lei Xu, Leyi Xia, Mingchuan Zhang, Minghua Zhang, Minghui Tang, Meng Li, Miaojun Wang, Mingming Li, Ning Tian, Panpan Huang, Peng Zhang, Qiancheng Wang, Qinyu Chen, Qiushi Du, Ruiqi Ge, Ruisong
  Zhang, Ruizhe Pan, Runji Wang, R.~J. Chen, R.~L. Jin, Ruyi Chen, Shanghao Lu, Shangyan Zhou, Shanhuang Chen, Shengfeng Ye, Shiyu Wang, Shuiping Yu, Shunfeng Zhou, Shuting Pan, S.~S. Li, Shuang Zhou, Shaoqing Wu, Shengfeng Ye, Tao Yun, Tian Pei, Tianyu Sun, T.~Wang, Wangding Zeng, Wanjia Zhao, Wen Liu, Wenfeng Liang, Wenjun Gao, Wenqin Yu, Wentao Zhang, W.~L. Xiao, Wei An, Xiaodong Liu, Xiaohan Wang, Xiaokang Chen, Xiaotao Nie, Xin Cheng, Xin Liu, Xin Xie, Xingchao Liu, Xinyu Yang, Xinyuan Li, Xuecheng Su, Xuheng Lin, X.~Q. Li, Xiangyue Jin, Xiaojin Shen, Xiaosha Chen, Xiaowen Sun, Xiaoxiang Wang, Xinnan Song, Xinyi Zhou, Xianzu Wang, Xinxia Shan, Y.~K. Li, Y.~Q. Wang, Y.~X. Wei, Yang Zhang, Yanhong Xu, Yao Li, Yao Zhao, Yaofeng Sun, Yaohui Wang, Yi~Yu, Yichao Zhang, Yifan Shi, Yiliang Xiong, Ying He, Yishi Piao, Yisong Wang, Yixuan Tan, Yiyang Ma, Yiyuan Liu, Yongqiang Guo, Yuan Ou, Yuduan Wang, Yue Gong, Yuheng Zou, Yujia He, Yunfan Xiong, Yuxiang Luo, Yuxiang You, Yuxuan Liu, Yuyang Zhou, Y.~X. Zhu,
  Yanhong Xu, Yanping Huang, Yaohui Li, Yi~Zheng, Yuchen Zhu, Yunxian Ma, Ying Tang, Yukun Zha, Yuting Yan, Z.~Z. Ren, Zehui Ren, Zhangli Sha, Zhe Fu, Zhean Xu, Zhenda Xie, Zhengyan Zhang, Zhewen Hao, Zhicheng Ma, Zhigang Yan, Zhiyu Wu, Zihui Gu, Zijia Zhu, Zijun Liu, Zilin Li, Ziwei Xie, Ziyang Song, Zizheng Pan, Zhen Huang, Zhipeng Xu, Zhongyu Zhang, and Zhen Zhang.
\newblock Deepseek-r1: Incentivizing reasoning capability in llms via reinforcement learning, 2025.

\bibitem{dodge2021documenting}
Jesse Dodge, Maarten Sap, Ana Marasović, William Agnew, Gabriel Ilharco, Dirk Groeneveld, Margaret Mitchell, and Matt Gardner.
\newblock Documenting large webtext corpora: A case study on the colossal clean crawled corpus, 2021.

\bibitem{llama3}
Aaron Grattafiori, Abhimanyu Dubey, Abhinav Jauhri, Abhinav Pandey, Abhishek Kadian, Ahmad Al-Dahle, Aiesha Letman, Akhil Mathur, Alan Schelten, Alex Vaughan, Amy Yang, Angela Fan, Anirudh Goyal, Anthony Hartshorn, Aobo Yang, Archi Mitra, Archie Sravankumar, Artem Korenev, Arthur Hinsvark, Arun Rao, Aston Zhang, Aurelien Rodriguez, Austen Gregerson, Ava Spataru, Baptiste Roziere, Bethany Biron, Binh Tang, Bobbie Chern, Charlotte Caucheteux, Chaya Nayak, Chloe Bi, Chris Marra, Chris McConnell, Christian Keller, Christophe Touret, Chunyang Wu, Corinne Wong, Cristian~Canton Ferrer, Cyrus Nikolaidis, Damien Allonsius, Daniel Song, Danielle Pintz, Danny Livshits, Danny Wyatt, David Esiobu, Dhruv Choudhary, Dhruv Mahajan, Diego Garcia-Olano, Diego Perino, Dieuwke Hupkes, Egor Lakomkin, Ehab AlBadawy, Elina Lobanova, Emily Dinan, Eric~Michael Smith, Filip Radenovic, Francisco Guzmán, Frank Zhang, Gabriel Synnaeve, Gabrielle Lee, Georgia~Lewis Anderson, Govind Thattai, Graeme Nail, Gregoire Mialon, Guan Pang,
  Guillem Cucurell, Hailey Nguyen, Hannah Korevaar, Hu~Xu, Hugo Touvron, Iliyan Zarov, Imanol~Arrieta Ibarra, Isabel Kloumann, Ishan Misra, Ivan Evtimov, Jack Zhang, Jade Copet, Jaewon Lee, Jan Geffert, Jana Vranes, Jason Park, Jay Mahadeokar, Jeet Shah, Jelmer van~der Linde, Jennifer Billock, Jenny Hong, Jenya Lee, Jeremy Fu, Jianfeng Chi, Jianyu Huang, Jiawen Liu, Jie Wang, Jiecao Yu, Joanna Bitton, Joe Spisak, Jongsoo Park, Joseph Rocca, Joshua Johnstun, Joshua Saxe, Junteng Jia, Kalyan~Vasuden Alwala, Karthik Prasad, Kartikeya Upasani, Kate Plawiak, Ke~Li, Kenneth Heafield, Kevin Stone, Khalid El-Arini, Krithika Iyer, Kshitiz Malik, Kuenley Chiu, Kunal Bhalla, Kushal Lakhotia, Lauren Rantala-Yeary, Laurens van~der Maaten, Lawrence Chen, Liang Tan, Liz Jenkins, Louis Martin, Lovish Madaan, Lubo Malo, Lukas Blecher, Lukas Landzaat, Luke de~Oliveira, Madeline Muzzi, Mahesh Pasupuleti, Mannat Singh, Manohar Paluri, Marcin Kardas, Maria Tsimpoukelli, Mathew Oldham, Mathieu Rita, Maya Pavlova, Melanie Kambadur,
  Mike Lewis, Min Si, Mitesh~Kumar Singh, Mona Hassan, Naman Goyal, Narjes Torabi, Nikolay Bashlykov, Nikolay Bogoychev, Niladri Chatterji, Ning Zhang, Olivier Duchenne, Onur Çelebi, Patrick Alrassy, Pengchuan Zhang, Pengwei Li, Petar Vasic, Peter Weng, Prajjwal Bhargava, Pratik Dubal, Praveen Krishnan, Punit~Singh Koura, Puxin Xu, Qing He, Qingxiao Dong, Ragavan Srinivasan, Raj Ganapathy, Ramon Calderer, Ricardo~Silveira Cabral, Robert Stojnic, Roberta Raileanu, Rohan Maheswari, Rohit Girdhar, Rohit Patel, Romain Sauvestre, Ronnie Polidoro, Roshan Sumbaly, Ross Taylor, Ruan Silva, Rui Hou, Rui Wang, Saghar Hosseini, Sahana Chennabasappa, Sanjay Singh, Sean Bell, Seohyun~Sonia Kim, Sergey Edunov, Shaoliang Nie, Sharan Narang, Sharath Raparthy, Sheng Shen, Shengye Wan, Shruti Bhosale, Shun Zhang, Simon Vandenhende, Soumya Batra, Spencer Whitman, Sten Sootla, Stephane Collot, Suchin Gururangan, Sydney Borodinsky, Tamar Herman, Tara Fowler, Tarek Sheasha, Thomas Georgiou, Thomas Scialom, Tobias Speckbacher,
  Todor Mihaylov, Tong Xiao, Ujjwal Karn, Vedanuj Goswami, Vibhor Gupta, Vignesh Ramanathan, Viktor Kerkez, Vincent Gonguet, Virginie Do, Vish Vogeti, Vítor Albiero, Vladan Petrovic, Weiwei Chu, Wenhan Xiong, Wenyin Fu, Whitney Meers, Xavier Martinet, Xiaodong Wang, Xiaofang Wang, Xiaoqing~Ellen Tan, Xide Xia, Xinfeng Xie, Xuchao Jia, Xuewei Wang, Yaelle Goldschlag, Yashesh Gaur, Yasmine Babaei, Yi~Wen, Yiwen Song, Yuchen Zhang, Yue Li, Yuning Mao, Zacharie~Delpierre Coudert, Zheng Yan, Zhengxing Chen, Zoe Papakipos, Aaditya Singh, Aayushi Srivastava, Abha Jain, Adam Kelsey, Adam Shajnfeld, Adithya Gangidi, Adolfo Victoria, Ahuva Goldstand, Ajay Menon, Ajay Sharma, Alex Boesenberg, Alexei Baevski, Allie Feinstein, Amanda Kallet, Amit Sangani, Amos Teo, Anam Yunus, Andrei Lupu, Andres Alvarado, Andrew Caples, Andrew Gu, Andrew Ho, Andrew Poulton, Andrew Ryan, Ankit Ramchandani, Annie Dong, Annie Franco, Anuj Goyal, Aparajita Saraf, Arkabandhu Chowdhury, Ashley Gabriel, Ashwin Bharambe, Assaf Eisenman, Azadeh
  Yazdan, Beau James, Ben Maurer, Benjamin Leonhardi, Bernie Huang, Beth Loyd, Beto~De Paola, Bhargavi Paranjape, Bing Liu, Bo~Wu, Boyu Ni, Braden Hancock, Bram Wasti, Brandon Spence, Brani Stojkovic, Brian Gamido, Britt Montalvo, Carl Parker, Carly Burton, Catalina Mejia, Ce~Liu, Changhan Wang, Changkyu Kim, Chao Zhou, Chester Hu, Ching-Hsiang Chu, Chris Cai, Chris Tindal, Christoph Feichtenhofer, Cynthia Gao, Damon Civin, Dana Beaty, Daniel Kreymer, Daniel Li, David Adkins, David Xu, Davide Testuggine, Delia David, Devi Parikh, Diana Liskovich, Didem Foss, Dingkang Wang, Duc Le, Dustin Holland, Edward Dowling, Eissa Jamil, Elaine Montgomery, Eleonora Presani, Emily Hahn, Emily Wood, Eric-Tuan Le, Erik Brinkman, Esteban Arcaute, Evan Dunbar, Evan Smothers, Fei Sun, Felix Kreuk, Feng Tian, Filippos Kokkinos, Firat Ozgenel, Francesco Caggioni, Frank Kanayet, Frank Seide, Gabriela~Medina Florez, Gabriella Schwarz, Gada Badeer, Georgia Swee, Gil Halpern, Grant Herman, Grigory Sizov, Guangyi, Zhang, Guna
  Lakshminarayanan, Hakan Inan, Hamid Shojanazeri, Han Zou, Hannah Wang, Hanwen Zha, Haroun Habeeb, Harrison Rudolph, Helen Suk, Henry Aspegren, Hunter Goldman, Hongyuan Zhan, Ibrahim Damlaj, Igor Molybog, Igor Tufanov, Ilias Leontiadis, Irina-Elena Veliche, Itai Gat, Jake Weissman, James Geboski, James Kohli, Janice Lam, Japhet Asher, Jean-Baptiste Gaya, Jeff Marcus, Jeff Tang, Jennifer Chan, Jenny Zhen, Jeremy Reizenstein, Jeremy Teboul, Jessica Zhong, Jian Jin, Jingyi Yang, Joe Cummings, Jon Carvill, Jon Shepard, Jonathan McPhie, Jonathan Torres, Josh Ginsburg, Junjie Wang, Kai Wu, Kam~Hou U, Karan Saxena, Kartikay Khandelwal, Katayoun Zand, Kathy Matosich, Kaushik Veeraraghavan, Kelly Michelena, Keqian Li, Kiran Jagadeesh, Kun Huang, Kunal Chawla, Kyle Huang, Lailin Chen, Lakshya Garg, Lavender A, Leandro Silva, Lee Bell, Lei Zhang, Liangpeng Guo, Licheng Yu, Liron Moshkovich, Luca Wehrstedt, Madian Khabsa, Manav Avalani, Manish Bhatt, Martynas Mankus, Matan Hasson, Matthew Lennie, Matthias Reso, Maxim
  Groshev, Maxim Naumov, Maya Lathi, Meghan Keneally, Miao Liu, Michael~L. Seltzer, Michal Valko, Michelle Restrepo, Mihir Patel, Mik Vyatskov, Mikayel Samvelyan, Mike Clark, Mike Macey, Mike Wang, Miquel~Jubert Hermoso, Mo~Metanat, Mohammad Rastegari, Munish Bansal, Nandhini Santhanam, Natascha Parks, Natasha White, Navyata Bawa, Nayan Singhal, Nick Egebo, Nicolas Usunier, Nikhil Mehta, Nikolay~Pavlovich Laptev, Ning Dong, Norman Cheng, Oleg Chernoguz, Olivia Hart, Omkar Salpekar, Ozlem Kalinli, Parkin Kent, Parth Parekh, Paul Saab, Pavan Balaji, Pedro Rittner, Philip Bontrager, Pierre Roux, Piotr Dollar, Polina Zvyagina, Prashant Ratanchandani, Pritish Yuvraj, Qian Liang, Rachad Alao, Rachel Rodriguez, Rafi Ayub, Raghotham Murthy, Raghu Nayani, Rahul Mitra, Rangaprabhu Parthasarathy, Raymond Li, Rebekkah Hogan, Robin Battey, Rocky Wang, Russ Howes, Ruty Rinott, Sachin Mehta, Sachin Siby, Sai~Jayesh Bondu, Samyak Datta, Sara Chugh, Sara Hunt, Sargun Dhillon, Sasha Sidorov, Satadru Pan, Saurabh Mahajan,
  Saurabh Verma, Seiji Yamamoto, Sharadh Ramaswamy, Shaun Lindsay, Shaun Lindsay, Sheng Feng, Shenghao Lin, Shengxin~Cindy Zha, Shishir Patil, Shiva Shankar, Shuqiang Zhang, Shuqiang Zhang, Sinong Wang, Sneha Agarwal, Soji Sajuyigbe, Soumith Chintala, Stephanie Max, Stephen Chen, Steve Kehoe, Steve Satterfield, Sudarshan Govindaprasad, Sumit Gupta, Summer Deng, Sungmin Cho, Sunny Virk, Suraj Subramanian, Sy~Choudhury, Sydney Goldman, Tal Remez, Tamar Glaser, Tamara Best, Thilo Koehler, Thomas Robinson, Tianhe Li, Tianjun Zhang, Tim Matthews, Timothy Chou, Tzook Shaked, Varun Vontimitta, Victoria Ajayi, Victoria Montanez, Vijai Mohan, Vinay~Satish Kumar, Vishal Mangla, Vlad Ionescu, Vlad Poenaru, Vlad~Tiberiu Mihailescu, Vladimir Ivanov, Wei Li, Wenchen Wang, Wenwen Jiang, Wes Bouaziz, Will Constable, Xiaocheng Tang, Xiaojian Wu, Xiaolan Wang, Xilun Wu, Xinbo Gao, Yaniv Kleinman, Yanjun Chen, Ye~Hu, Ye~Jia, Ye~Qi, Yenda Li, Yilin Zhang, Ying Zhang, Yossi Adi, Youngjin Nam, Yu, Wang, Yu~Zhao, Yuchen Hao, Yundi
  Qian, Yunlu Li, Yuzi He, Zach Rait, Zachary DeVito, Zef Rosnbrick, Zhaoduo Wen, Zhenyu Yang, Zhiwei Zhao, and Zhiyu Ma.
\newblock The llama 3 herd of models, 2024.

\bibitem{gururangan-etal-2022-whose}
Suchin Gururangan, Dallas Card, Sarah Dreier, Emily Gade, Leroy Wang, Zeyu Wang, Luke Zettlemoyer, and Noah~A. Smith.
\newblock Whose language counts as high quality? measuring language ideologies in text data selection.
\newblock In Yoav Goldberg, Zornitsa Kozareva, and Yue Zhang, editors, {\em Proceedings of the 2022 Conference on Empirical Methods in Natural Language Processing}, pages 2562--2580, Abu Dhabi, United Arab Emirates, December 2022. Association for Computational Linguistics.

\bibitem{TALE-EP}
Tingxu Han, Zhenting Wang, Chunrong Fang, Shiyu Zhao, Shiqing Ma, and Zhenyu Chen.
\newblock Token-budget-aware llm reasoning.
\newblock {\em arXiv preprint arXiv:2412.18547}, 2024.

\bibitem{COCONUT}
Shibo Hao, Sainbayar Sukhbaatar, DiJia Su, Xian Li, Zhiting Hu, Jason Weston, and Yuandong Tian.
\newblock Training large language models to reason in a continuous latent space.
\newblock {\em arXiv preprint arXiv:2412.06769}, 2024.

\bibitem{DeepMath-103K}
Zhiwei He, Tian Liang, Jiahao Xu, Qiuzhi Liu, Xingyu Chen, Yue Wang, Linfeng Song, Dian Yu, Zhenwen Liang, Wenxuan Wang, Zhuosheng Zhang, Rui Wang, Zhaopeng Tu, Haitao Mi, and Dong Yu.
\newblock Deepmath-103k: A large-scale, challenging, decontaminated, and verifiable mathematical dataset for advancing reasoning, 2025.

\bibitem{hendrycks2021measuring}
Dan Hendrycks, Collin Burns, Saurav Kadavath, Akul Arora, Steven Basart, Eric Tang, Dawn Song, and Jacob Steinhardt.
\newblock Measuring mathematical problem solving with the math dataset.
\newblock {\em arXiv preprint arXiv:2103.03874}, 2021.

\bibitem{hernandez2022scaling}
Danny Hernandez, Tom Brown, Tom Conerly, Nova DasSarma, Dawn Drain, Sheer El-Showk, Nelson Elhage, Zac Hatfield-Dodds, Tom Henighan, Tristan Hume, Scott Johnston, Ben Mann, Chris Olah, Catherine Olsson, Dario Amodei, Nicholas Joseph, Jared Kaplan, and Sam McCandlish.
\newblock Scaling laws and interpretability of learning from repeated data, 2022.

\bibitem{thinkpruner}
Bairu Hou, Yang Zhang, Jiabao Ji, Yujian Liu, Kaizhi Qian, Jacob Andreas, and Shiyu Chang.
\newblock Thinkprune: Pruning long chain-of-thought of llms via reinforcement learning, 2025.

\bibitem{thinkerprune}
Bairu Hou, Yang Zhang, Jiabao Ji, Yujian Liu, Kaizhi Qian, Jacob Andreas, and Shiyu Chang.
\newblock Thinkprune: Pruning long chain-of-thought of llms via reinforcement learning, 2025.

\bibitem{huang2024effi}
Dong Huang, Guangtao Zeng, Jianbo Dai, Meng Luo, Han Weng, Yuhao Qing, Heming Cui, Zhijiang Guo, and Jie~M Zhang.
\newblock Effi-code: Unleashing code efficiency in language models.
\newblock {\em arXiv preprint arXiv:2410.10209}, 2024.

\bibitem{jiang2025drp}
Yuxuan Jiang, Dawei Li, and Frank Ferraro.
\newblock Drp: Distilled reasoning pruning with skill-aware step decomposition for efficient large reasoning models.
\newblock {\em arXiv preprint arXiv:2505.13975}, 2025.

\bibitem{C3oT}
Yu~Kang, Xianghui Sun, Liangyu Chen, and Wei Zou.
\newblock C3ot: Generating shorter chain-of-thought without compromising effectiveness.
\newblock {\em arXiv preprint arXiv:2412.11664}, 2024.

\bibitem{kumar2025overthink}
Abhinav Kumar, Jaechul Roh, Ali Naseh, Marzena Karpinska, Mohit Iyyer, Amir Houmansadr, and Eugene Bagdasarian.
\newblock Overthink: Slowdown attacks on reasoning llms.
\newblock {\em arXiv preprint arXiv:2502.02542}, 2025.

\bibitem{bigscience}
Hugo Laurençon, Lucile Saulnier, Thomas Wang, Christopher Akiki, Albert~Villanova del Moral, Teven~Le Scao, Leandro~Von Werra, Chenghao Mou, Eduardo~González Ponferrada, Huu Nguyen, et~al.
\newblock The bigscience roots corpus: A 1.6tb composite multilingual dataset, 2023.

\bibitem{ConciseCoT}
Ayeong Lee, Ethan Che, and Tianyi Peng.
\newblock How well do llms compress their own chain-of-thought? a token complexity approach, 2025.

\bibitem{lee2022deduplicating}
Katherine Lee, Daphne Ippolito, Andrew Nystrom, Chiyuan Zhang, Douglas Eck, Chris Callison-Burch, and Nicholas Carlini.
\newblock Deduplicating training data makes language models better, 2022.

\bibitem{li_system_2_reason}
Zhong-Zhi Li, Duzhen Zhang, Ming-Liang Zhang, Jiaxin Zhang, Zengyan Liu, Yuxuan Yao, Haotian Xu, Junhao Zheng, Pei-Jie Wang, Xiuyi Chen, Yingying Zhang, Fei Yin, Jiahua Dong, Zhiwei Li, Bao-Long Bi, Ling-Rui Mei, Junfeng Fang, Zhijiang Guo, Le~Song, and Cheng-Lin Liu.
\newblock From system 1 to system 2: A survey of reasoning large language models, 2025.

\bibitem{liu2024datasets}
Yang Liu, Jiahuan Cao, Chongyu Liu, Kai Ding, and Lianwen Jin.
\newblock Datasets for large language models: A comprehensive survey, 2024.

\bibitem{longpre2023pretrainers}
Shayne Longpre, Gregory Yauney, Emily Reif, Katherine Lee, Adam Roberts, Barret Zoph, Denny Zhou, Jason Wei, Kevin Robinson, David Mimno, and Daphne Ippolito.
\newblock A pretrainer's guide to training data: Measuring the effects of data age, domain coverage, quality, \& toxicity, 2023.

\bibitem{O1-Pruner}
Haotian Luo, Li~Shen, Haiying He, Yibo Wang, Shiwei Liu, Wei Li, Naiqiang Tan, Xiaochun Cao, and Dacheng Tao.
\newblock O1-pruner: Length-harmonizing fine-tuning for o1-like reasoning pruning, 2025.

\bibitem{o1_pruner}
Haotian Luo, Li~Shen, Haiying He, Yibo Wang, Shiwei Liu, Wei Li, Naiqiang Tan, Xiaochun Cao, and Dacheng Tao.
\newblock {O1-Pruner: Length-Harmonizing Fine-Tuning for O1-Like Reasoning Pruning}.
\newblock {\em arXiv preprint arXiv:2501.12570}, 2025.

\bibitem{CoT-Valve}
Xinyin Ma, Guangnian Wan, Runpeng Yu, Gongfan Fang, and Xinchao Wang.
\newblock Cot-valve: Length-compressible chain-of-thought tuning, 2025.

\bibitem{SimPO}
Yu~Meng, Mengzhou Xia, and Danqi Chen.
\newblock Simpo: Simple preference optimization with a reference-free reward, 2024.

\bibitem{miao2021diverse}
Shen-Yun Miao, Chao-Chun Liang, and Keh-Yih Su.
\newblock A diverse corpus for evaluating and developing english math word problem solvers.
\newblock {\em arXiv preprint arXiv:2106.15772}, 2021.

\bibitem{muennighoff2023scaling}
Niklas Muennighoff, Alexander~M. Rush, Boaz Barak, Teven~Le Scao, Aleksandra Piktus, Nouamane Tazi, Sampo Pyysalo, Thomas Wolf, and Colin Raffel.
\newblock Scaling data-constrained language models, 2023.

\bibitem{s1}
Niklas Muennighoff, Zitong Yang, Weijia Shi, Xiang~Lisa Li, Li~Fei-Fei, Hannaneh Hajishirzi, Luke Zettlemoyer, Percy Liang, Emmanuel Candès, and Tatsunori Hashimoto.
\newblock s1: Simple test-time scaling, 2025.

\bibitem{RouteLLM}
Isaac Ong, Amjad Almahairi, Vincent Wu, Wei-Lin Chiang, Tianhao Wu, Joseph~E. Gonzalez, M~Waleed Kadous, and Ion Stoica.
\newblock Routellm: Learning to route llms with preference data, 2025.

\bibitem{GPT4o}
OpenAI, Josh Achiam, Steven Adler, Sandhini Agarwal, Lama Ahmad, Ilge Akkaya, Florencia~Leoni Aleman, Diogo Almeida, Janko Altenschmidt, Sam Altman, Shyamal Anadkat, Red Avila, Igor Babuschkin, Suchir Balaji, Valerie Balcom, Paul Baltescu, Haiming Bao, Mohammad Bavarian, Jeff Belgum, Irwan Bello, Jake Berdine, Gabriel Bernadett-Shapiro, Christopher Berner, Lenny Bogdonoff, Oleg Boiko, Madelaine Boyd, Anna-Luisa Brakman, Greg Brockman, Tim Brooks, Miles Brundage, Kevin Button, Trevor Cai, Rosie Campbell, Andrew Cann, Brittany Carey, Chelsea Carlson, Rory Carmichael, Brooke Chan, Che Chang, Fotis Chantzis, Derek Chen, Sully Chen, Ruby Chen, Jason Chen, Mark Chen, Ben Chess, Chester Cho, Casey Chu, Hyung~Won Chung, Dave Cummings, Jeremiah Currier, Yunxing Dai, Cory Decareaux, Thomas Degry, Noah Deutsch, Damien Deville, Arka Dhar, David Dohan, Steve Dowling, Sheila Dunning, Adrien Ecoffet, Atty Eleti, Tyna Eloundou, David Farhi, Liam Fedus, Niko Felix, Simón~Posada Fishman, Juston Forte, Isabella Fulford, Leo
  Gao, Elie Georges, Christian Gibson, Vik Goel, Tarun Gogineni, Gabriel Goh, Rapha Gontijo-Lopes, Jonathan Gordon, Morgan Grafstein, Scott Gray, Ryan Greene, Joshua Gross, Shixiang~Shane Gu, Yufei Guo, Chris Hallacy, Jesse Han, Jeff Harris, Yuchen He, Mike Heaton, Johannes Heidecke, Chris Hesse, Alan Hickey, Wade Hickey, Peter Hoeschele, Brandon Houghton, Kenny Hsu, Shengli Hu, Xin Hu, Joost Huizinga, Shantanu Jain, Shawn Jain, Joanne Jang, Angela Jiang, Roger Jiang, Haozhun Jin, Denny Jin, Shino Jomoto, Billie Jonn, Heewoo Jun, Tomer Kaftan, Łukasz Kaiser, Ali Kamali, Ingmar Kanitscheider, Nitish~Shirish Keskar, Tabarak Khan, Logan Kilpatrick, Jong~Wook Kim, Christina Kim, Yongjik Kim, Jan~Hendrik Kirchner, Jamie Kiros, Matt Knight, Daniel Kokotajlo, Łukasz Kondraciuk, Andrew Kondrich, Aris Konstantinidis, Kyle Kosic, Gretchen Krueger, Vishal Kuo, Michael Lampe, Ikai Lan, Teddy Lee, Jan Leike, Jade Leung, Daniel Levy, Chak~Ming Li, Rachel Lim, Molly Lin, Stephanie Lin, Mateusz Litwin, Theresa Lopez, Ryan
  Lowe, Patricia Lue, Anna Makanju, Kim Malfacini, Sam Manning, Todor Markov, Yaniv Markovski, Bianca Martin, Katie Mayer, Andrew Mayne, Bob McGrew, Scott~Mayer McKinney, Christine McLeavey, Paul McMillan, Jake McNeil, David Medina, Aalok Mehta, Jacob Menick, Luke Metz, Andrey Mishchenko, Pamela Mishkin, Vinnie Monaco, Evan Morikawa, Daniel Mossing, Tong Mu, Mira Murati, Oleg Murk, David Mély, Ashvin Nair, Reiichiro Nakano, Rajeev Nayak, Arvind Neelakantan, Richard Ngo, Hyeonwoo Noh, Long Ouyang, Cullen O'Keefe, Jakub Pachocki, Alex Paino, Joe Palermo, Ashley Pantuliano, Giambattista Parascandolo, Joel Parish, Emy Parparita, Alex Passos, Mikhail Pavlov, Andrew Peng, Adam Perelman, Filipe de~Avila Belbute~Peres, Michael Petrov, Henrique~Ponde de~Oliveira~Pinto, Michael, Pokorny, Michelle Pokrass, Vitchyr~H. Pong, Tolly Powell, Alethea Power, Boris Power, Elizabeth Proehl, Raul Puri, Alec Radford, Jack Rae, Aditya Ramesh, Cameron Raymond, Francis Real, Kendra Rimbach, Carl Ross, Bob Rotsted, Henri Roussez,
  Nick Ryder, Mario Saltarelli, Ted Sanders, Shibani Santurkar, Girish Sastry, Heather Schmidt, David Schnurr, John Schulman, Daniel Selsam, Kyla Sheppard, Toki Sherbakov, Jessica Shieh, Sarah Shoker, Pranav Shyam, Szymon Sidor, Eric Sigler, Maddie Simens, Jordan Sitkin, Katarina Slama, Ian Sohl, Benjamin Sokolowsky, Yang Song, Natalie Staudacher, Felipe~Petroski Such, Natalie Summers, Ilya Sutskever, Jie Tang, Nikolas Tezak, Madeleine~B. Thompson, Phil Tillet, Amin Tootoonchian, Elizabeth Tseng, Preston Tuggle, Nick Turley, Jerry Tworek, Juan Felipe~Cerón Uribe, Andrea Vallone, Arun Vijayvergiya, Chelsea Voss, Carroll Wainwright, Justin~Jay Wang, Alvin Wang, Ben Wang, Jonathan Ward, Jason Wei, CJ~Weinmann, Akila Welihinda, Peter Welinder, Jiayi Weng, Lilian Weng, Matt Wiethoff, Dave Willner, Clemens Winter, Samuel Wolrich, Hannah Wong, Lauren Workman, Sherwin Wu, Jeff Wu, Michael Wu, Kai Xiao, Tao Xu, Sarah Yoo, Kevin Yu, Qiming Yuan, Wojciech Zaremba, Rowan Zellers, Chong Zhang, Marvin Zhang, Shengjia
  Zhao, Tianhao Zheng, Juntang Zhuang, William Zhuk, and Barret Zoph.
\newblock Gpt-4 technical report, 2024.

\bibitem{rae2022scaling}
Jack~W. Rae, Sebastian Borgeaud, Trevor Cai, Katie Millican, Jordan Hoffmann, Francis Song, John Aslanides, Sarah Henderson, Roman Ring, Susannah Young, et~al.
\newblock Scaling language models: Methods, analysis \& insights from training gopher, 2022.

\bibitem{raffel2023exploring}
Colin Raffel, Noam Shazeer, Adam Roberts, Katherine Lee, Sharan Narang, Michael Matena, Yanqi Zhou, Wei Li, and Peter~J. Liu.
\newblock Exploring the limits of transfer learning with a unified text-to-text transformer, 2023.

\bibitem{Heima}
Xuan Shen, Yizhou Wang, Xiangxi Shi, Yanzhi Wang, Pu~Zhao, and Jiuxiang Gu.
\newblock Efficient reasoning with hidden thinking.
\newblock {\em arXiv preprint arXiv:2501.19201}, 2025.

\bibitem{team2025kimi}
Kimi Team, Angang Du, Bofei Gao, Bowei Xing, Changjiu Jiang, Cheng Chen, Cheng Li, Chenjun Xiao, Chenzhuang Du, Chonghua Liao, et~al.
\newblock {Kimi k1. 5: Scaling Reinforcement Learning with LLMs}.
\newblock {\em arXiv preprint arXiv:2501.12599}, 2025.

\bibitem{wang2024explore}
Shu Wang, Lei Ji, Renxi Wang, Wenxiao Zhao, Haokun Liu, Yifan Hou, and Ying~Nian Wu.
\newblock Explore the reasoning capability of llms in the chess testbed.
\newblock {\em arXiv preprint arXiv:2411.06655}, 2024.

\bibitem{bpo}
Sizhe Wang, Yongqi Tong, Hengyuan Zhang, Dawei Li, Xin Zhang, and Tianlong Chen.
\newblock Bpo: Towards balanced preference optimization between knowledge breadth and depth in alignment.
\newblock {\em arXiv preprint arXiv:2411.10914}, 2024.

\bibitem{Merging-Long2Short}
Han Wu, Yuxuan Yao, Shuqi Liu, Zehua Liu, Xiaojin Fu, Xiongwei Han, Xing Li, Hui-Ling Zhen, Tao Zhong, and Mingxuan Yuan.
\newblock Unlocking efficient long-to-short llm reasoning with model merging, 2025.

\bibitem{OS-ATLAS}
Zhiyong Wu, Zhenyu Wu, Fangzhi Xu, Yian Wang, Qiushi Sun, Chengyou Jia, Kanzhi Cheng, Zichen Ding, Liheng Chen, Paul~Pu Liang, and Yu~Qiao.
\newblock Os-atlas: A foundation action model for generalist gui agents, 2024.

\bibitem{TokenSkip}
Heming Xia, Yongqi Li, Chak~Tou Leong, Wenjie Wang, and Wenjie Li.
\newblock Tokenskip: Controllable chain-of-thought compression in llms.
\newblock {\em arXiv preprint arXiv:2502.12067}, 2025.

\bibitem{xu2021detoxifying}
Albert Xu, Eshaan Pathak, Eric Wallace, Suchin Gururangan, Maarten Sap, and Dan Klein.
\newblock Detoxifying language models risks marginalizing minority voices, 2021.

\bibitem{RedStar}
Haotian Xu, Xing Wu, Weinong Wang, Zhongzhi Li, Da~Zheng, Boyuan Chen, Yi~Hu, Shijia Kang, Jiaming Ji, Yingying Zhang, et~al.
\newblock {RedStar: Does Scaling Long-CoT Data Unlock Better Slow-Reasoning Systems?}
\newblock {\em arXiv preprint arXiv:2501.11284}, 2025.

\bibitem{CoD}
Silei Xu, Wenhao Xie, Lingxiao Zhao, and Pengcheng He.
\newblock Chain of draft: Thinking faster by writing less.
\newblock {\em arXiv preprint arXiv:2502.18600}, 2025.

\bibitem{TOPS}
Wenkai Yang, Shuming Ma, Yankai Lin, and Furu Wei.
\newblock Towards thinking-optimal scaling of test-time compute for llm reasoning, 2025.

\bibitem{yao2025activation}
Yuxuan Yao, Shuqi Liu, Zehua Liu, Qintong Li, Mingyang Liu, Xiongwei Han, Zhijiang Guo, Han Wu, and Linqi Song.
\newblock Activation-guided consensus merging for large language models.
\newblock {\em arXiv preprint arXiv:2505.14009}, 2025.

\bibitem{Reasoning-in-Flux}
Zhangyue Yin, Qiushi Sun, Qipeng Guo, Zhiyuan Zeng, Xiaonan Li, Junqi Dai, Qinyuan Cheng, Xuanjing Huang, and Xipeng Qiu.
\newblock Reasoning in flux: Enhancing large language models reasoning through uncertainty-aware adaptive guidance.
\newblock In Lun-Wei Ku, Andre Martins, and Vivek Srikumar, editors, {\em Proceedings of the 62nd Annual Meeting of the Association for Computational Linguistics (Volume 1: Long Papers)}, pages 2401--2416, Bangkok, Thailand, August 2024. Association for Computational Linguistics.

\bibitem{yu2024distilling21}
Ping Yu, Jing Xu, Jason Weston, and Ilia Kulikov.
\newblock Distilling system 2 into system 1, 2024.

\bibitem{chain-of-reasoning}
Yiyao Yu, Yuxiang Zhang, Dongdong Zhang, Xiao Liang, Hengyuan Zhang, Xingxing Zhang, Ziyi Yang, Mahmoud Khademi, Hany Awadalla, Junjie Wang, et~al.
\newblock Chain-of-reasoning: Towards unified mathematical reasoning in large language models via a multi-paradigm perspective.
\newblock {\em arXiv preprint arXiv:2501.11110}, 2025.

\bibitem{zhang_balancing}
Hengyuan Zhang, Yanru Wu, Dawei Li, Zacc Yang, Rui Zhao, Yong Jiang, and Fei Tan.
\newblock Balancing speciality and versatility: a coarse to fine framework for supervised fine-tuning large language model.
\newblock {\em arXiv e-prints}, pages arXiv--2404, 2024.

\bibitem{SPRO}
Xiaojiang Zhang, Jinghui Wang, Zifei Cheng, Wenhao Zhuang, Zheng Lin, Minglei Zhang, Shaojie Wang, Yinghan Cui, Chao Wang, Junyi Peng, Shimiao Jiang, Shiqi Kuang, Shouyu Yin, Chaohang Wen, Haotian Zhang, Bin Chen, and Bing Yu.
\newblock Srpo: A cross-domain implementation of large-scale reinforcement learning on llm, 2025.

\end{thebibliography}

\newpage

\appendix

\clearpage
\section{Appendix}
\label{sec:appendix}
\subsection{Training Details}
Due to the need to evaluate accuracy and token count on a validation set every n steps, our validation set consists of 512 questions sampled from past questions in AIME-1983 to AIME-2023. The original ratio for shortcot and longcot is set to 0.5:0.5, with an evaluation interval of every 32 steps. The model is allowed to train for a total of 2000 steps, and the learning rate is set as a constant at 1e-5. For the 7B and 14B models, we conducted training on two 8-GPU (80GB) machines, with one 8-GPU machine performing vllm inference and the other performing training. Every n steps, parameter synchronization is executed using vllm's parameter sync function. For the 32B model, we performed SFT on five 8-GPU nodes and deployed the inference service on one node.

We ultimately select the checkpoint with the shortest token length among those whose accuracy on the validation set is no less than 30\% of that achieved by the original long CoT.
\subsection{Metrics}
\subsubsection{Compression Rate}
We provide more details on the compression rate in the main table, where the compression rate is defined as:
\begin{equation}
\text{C.R.} =   \text{Compression Rate} = max(\frac{\# \text{tokens}_{\text{original}} - \# \text{tokens}_{\text{current}}}{\# \text{tokens}_{\text{original}}}, 0)
\label{equ:compreesion_rate}
\end{equation}
\begin{equation}
\text{A.C.R.} = \frac{1}{N_{\text{benchmark}}}\sum_{i=0}^{N_{\text{benchmark}}}\text{C.R.}
\end{equation}
\subsubsection{Normalized Metric}
We report two normalized metrics to facilitate fair comparisons: Normalized Accuracy and Normalized Token Length. They are defined as follows:
\begin{equation}
\text{Normalized Accuracy} = \frac{\#\text{Acc}_{\text{current}}}{\#\text{Acc}_{\text{original}}}
\label{equ:normlized_acc}
\end{equation}
\begin{equation}
\text{Normalized Token} = \frac{\#\text{Token}_{\text{current}}}{\#\text{Token}_{\text{original}}}
\label{equ:normlized_token}
\end{equation}

\subsection{Data Construction Detail}
For long CoT, we use the prompt from dataset s1.1 ~\cite{s1}. Each sample is generated 8 times using the original model.
For short CoT, to avoid inconsistencies in the system prompt format, we adopt the short CoT construction method from AdaR1. We annotate 10 randomly selected questions from GSM8K using the instruct model, then fine-tune the long CoT model to overfit on them. For the GSM8K training set, we sample and retain only the examples with correct answers.
\subsection{Evaluation Detail} 
We use the DeepSeek-R1-Distill model and apply a temperature setting of 0.7, which is the primary recommendation in QwQ-Preview, for evaluating all models. All datasets are restricted to an 8K context window for output generation.
Meanwhile, considering the relatively small sizes of the AMC and AIME datasets, we sample 8 responses per question and compute the average.
\subsection{Evaluation Framework} 
We use \textit{skythought-eval}\footnote{\url{https://github.com/NovaSky-AI/SkyThought}} as the framework, which supports accelerating long CoT reasoning evaluation with vLLM. The version of vLLM we use is 0.6.3.\\


\subsection{Evaluation Dataset Detail}
We provide an overview of all datasets used in the following sections.
\begin{itemize}
    \item \textbf{ASDiv}: A  diverse simple English math word problem  corpus for evaluating the capability of various MWP solvers. It contains 2,305 MWPs that cover more text patterns and most problem types taught in elementary school. 
    \item \textbf{GSM8K}: A high-quality benchmark comprising 8,500 human-written grade school math word problems that require multi-step reasoning and basic arithmetic, each labeled with a natural language solution and verified answer. The 1,319-question test set emphasizes sequential reasoning and is primarily solvable by upper-grade elementary school students.
    \item \textbf{MATH500}: a challenging benchmark of 500 high school competition-level problems spanning seven subjects, including Algebra, Geometry, Number Theory, and Precalculus. Each problem is presented in natural language with LaTeX-formatted notation, offering a strong measure of mathematical reasoning and generalization across diverse topics.
    \item \textbf{AIME2024}: a dataset containing 30 problems from the 2024 American Invitational Mathematics Examination (AIME), a prestigious high school mathematics competition for top-performing students. Each problem is designed to require deep mathematical insight, multi-step reasoning, and precise problem-solving skills.
    \item \textbf{AMC}: The AMC dataset consists of all 83 problems from AMC12 2022 and AMC12 2023, extracted from the AoPS wiki page. We used a subset of this data containing 40 problems.
    \item \textbf{MinervaMath}: MinervaMath is a high-difficulty math problem dataset containing 272 challenging problems.
\end{itemize}
\subsection{Reproduce Details}
\noindent \textbf{ConciseCoT \& TALE-EP}
For the prompt-based baseline, we list the prompts used in Prompt \ref{tab:evaluation_prompt_figure}. \\
\textbf{OverThink} 
For the MATH12K dataset, we sample each problem 8 times. The shortest correct sample is selected as the chosen sample, and the longest sample is selected as the rejected sample. The model is trained for 1 epoch. \\
\textbf{ThinkPruner} 
In our reproduction, we use the competition-level training data provided in the original paper and train the model for 10 epochs with a learning rate of 1e-6. The maximum response length is set to 4096 tokens. We follow their early stopping strategy to select the optimal checkpoint for evaluation. \\
\textbf{CoT-Valve} 
Since CoT-Valve does not report performance on all datasets, we reproduced the results using the public datasets released by CoT-Valve. We followed the training settings officially reported in the paper, using LoRA=2 to fine-tune all models. The dataset version used is Mix-Chain-Z-GSM8K. All models were fine-tuned for 5 epochs on 8 GPUs with 80GB of memory each. \\

\begin{figure*}[t]
\centering
\small
\begin{tcolorbox}[colback=cyan!5!white, 
                  colframe=cyan!80!black, 
                  width=0.99\textwidth, 
                  arc=3mm, 
                  title={Evaluation Prompt on Dataset},
                  auto outer arc,
                  ]

\vspace{0.5em}
\textbf{=== EVALUATION PROMPT FOR GSM8K ===} \\
<|begin\_of\_sentence|>Please reason step by step, and put your final answer within 'boxed'.\\<|User|>{query}<|Assistant|>Given the following problem, reason and give a final answer to the problem. \\ Problem: \{question\}\\ Your response should end with \"The final answer is [answer]\" where [answer] is the response to the problem. <think> \\ \\
\textbf{=== EVALUATION PROMPT FOR MATH500 ===} \\
<|begin\_of\_sentence|>Please reason step by step, and put your final answer within '\\boxed'. <|User|>\{query\}<|Assistant|>Return your final response within 'boxed'. \{problem\}. <think> \\ \\
\textbf{=== EVALUATION PROMPT FOR AIME24 ===} \\
<|begin\_of\_sentence|>Please reason step by step, and put your final answer within '\\boxed'. <|User|>{query}<|Assistant|>Return your final response within '\\boxed'. \{problem\}. <think> \\ \\
\textbf{=== EVALUATION PROMPT FOR AMC ===} \\
<|begin\_of\_sentence|>Please reason step by step, and put your final answer within 'boxed'. <|User|>{query}<|Assistant|>Return your final response within 'boxed'. \{problem\}. <think> \\ \\
\textbf{=== EVALUATION PROMPT FOR MINERVAMATH ===} \\
<|begin\_of\_sentence|>Please reason step by step, and put your final answer within 'boxed'. <|User|>{query}<|Assistant|>Return your final response within 'boxed'. {problem}. <think> \\ 
\end{tcolorbox}
\caption{Evaluation Prompt for GSM8K, MATH500, AIME24, AMC, MinervaMath}
\label{tab:evaluation_prompt_figure}
\end{figure*}

\textbf{L1} 
In L1 reproduction on the 7B System-2 model, we utilize the \textit{L1-Exact} reward function and limit the token length to between 100 and 4,096 tokens, while setting the token difference penalization parameter $\alpha$ to 0.0003, as described in the paper. We follow their original prompt by appending "Think for $n_{token}$ tokens" to the end of the question. In inference, the token budget is set to the same number as the average tokens from our method across the evaluated benchmarks. \\

\section{Case Study}

\begin{figure*}
    \centering
    \includegraphics[width=1.0\linewidth]{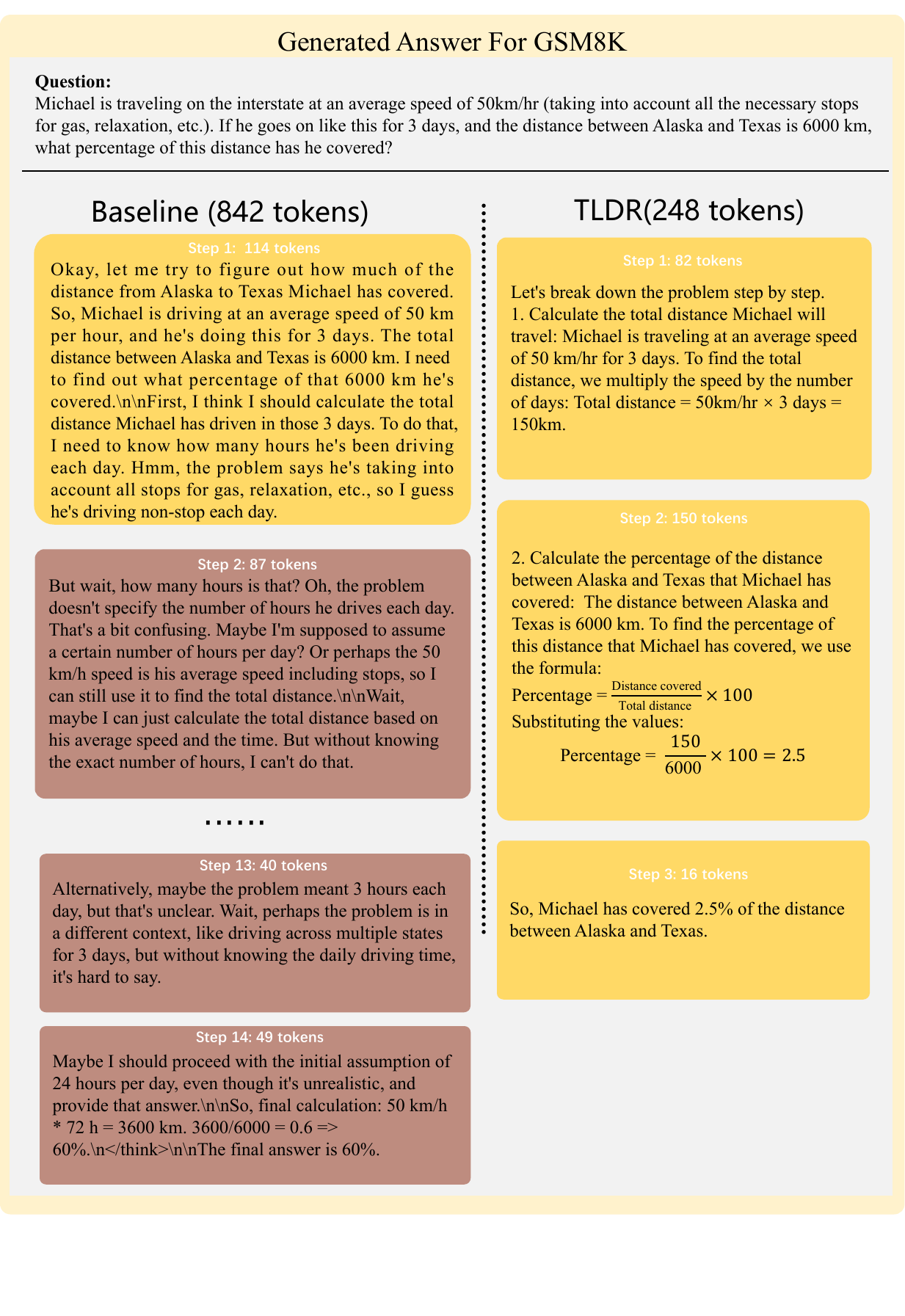}
    \vspace{-20mm}
    \caption{Comparison of Reasoning process on GSM8K: Baseline vs. TLDR.}
    \label{fig:gsm8k}
\end{figure*}

\begin{figure*}
    \centering
    \includegraphics[width=1.0\linewidth]{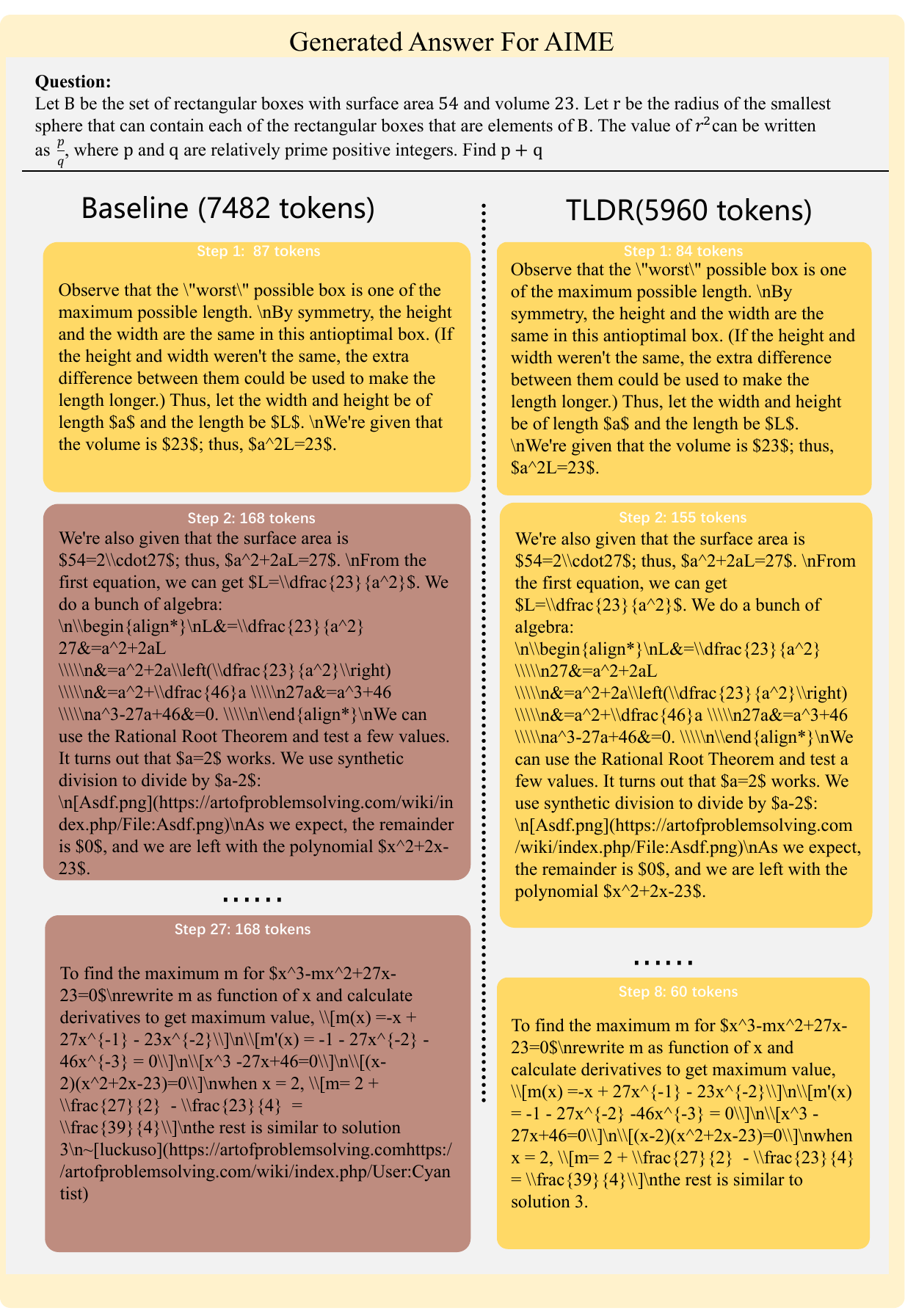}
    \vspace{-1mm}
    \caption{Comparison of Reasoning process on AIME: Baseline vs. TLDR.}
    \label{fig:aime}
\end{figure*}

\begin{figure*}
    \centering
    \includegraphics[width=1.0\linewidth]{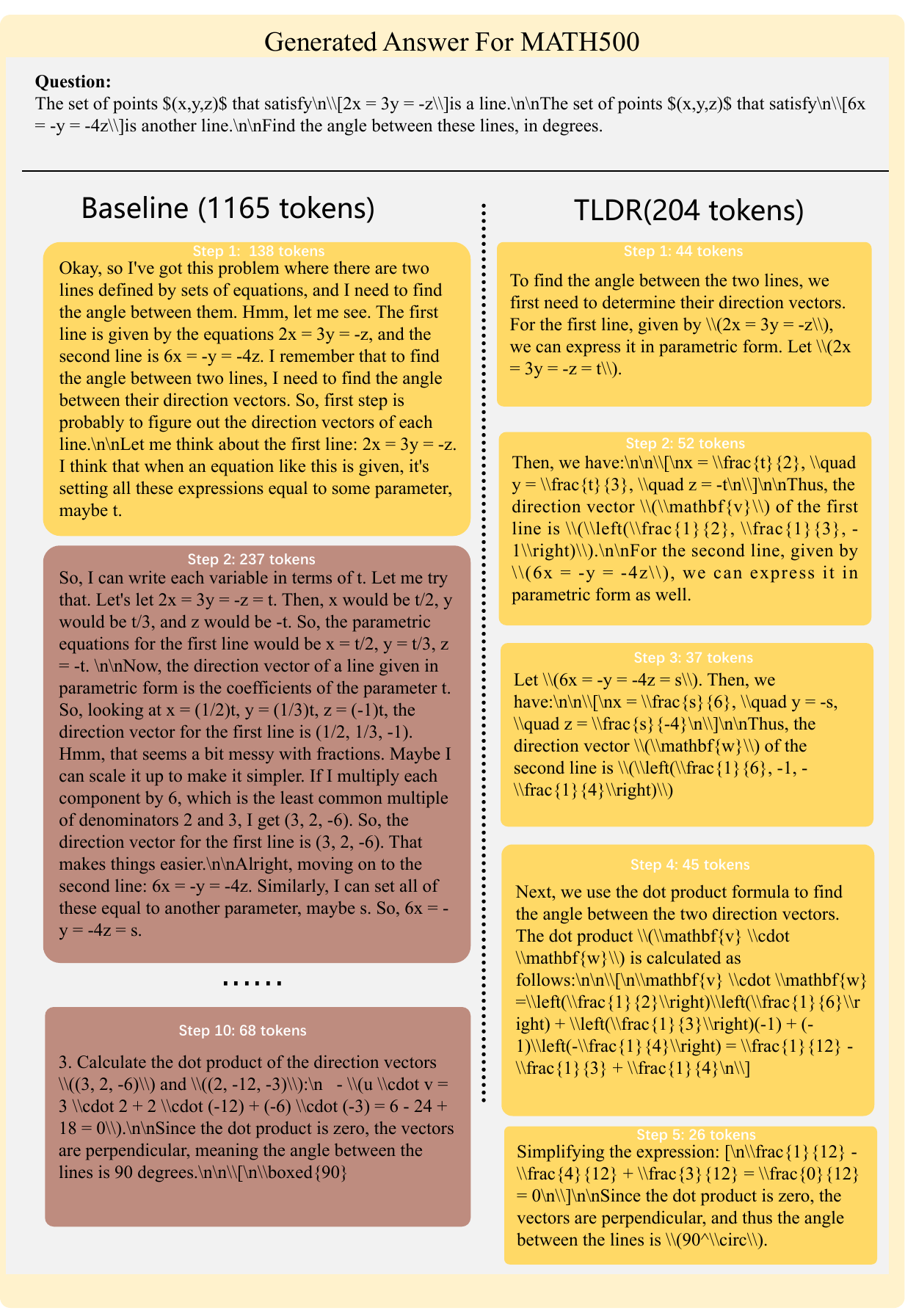}
    \vspace{-1mm}
    \caption{Comparison of Reasoning process on MATH500: Baseline vs. TLDR.}
    \label{fig:math500}
\end{figure*}

To better understand the behavioral differences between baseline and TLDR strategies, we conduct a qualitative analysis using the DeepSeek-R1-Distill-Qwen-7B model. Case studies are drawn from three representative math datasets: GSM8K, AIME, and MATH500. As shown in Figures~\ref{fig:gsm8k}–\ref{fig:math500}, the baseline model tends to generate verbose reasoning paths with redundant or speculative content. In contrast, TLDR produces significantly more concise outputs while maintaining correctness and logical structure. These examples demonstrate TLDR’s ability to suppress unnecessary reasoning tokens—such as exploratory or reflective phrases—leading to more efficient and focused reasoning processes.

\clearpage

\clearpage 
\end{document}